\pdfoutput=1

\documentclass[11pt]{article}

\usepackage[]{ACL2023}

\usepackage{times}
\usepackage{latexsym}

\usepackage[T1]{fontenc}

\usepackage[russian,english]{babel}
\usepackage[utf8]{inputenc}

\usepackage{microtype}

\usepackage{inconsolata}

\usepackage{booktabs}
\usepackage{multirow}
\usepackage{tablefootnote}
\usepackage{graphicx}
\usepackage{array}  % For column shrinking

\usepackage{caption}
\usepackage{subcaption}

\title{Bilingual Rhetorical Structure Parsing with Large Parallel Annotations}

\author{Elena Chistova \\ FRC CSC RAS \quad ISP RAS \\
  \texttt{chistova@isa.ru} 
}

\begin{document}
\maketitle
\begin{abstract}

Discourse parsing is a crucial task in natural language processing that aims to reveal the higher-level relations in a text. Despite growing interest in cross-lingual discourse parsing, challenges persist due to limited parallel data and inconsistencies in the Rhetorical Structure Theory (RST) application across languages and corpora. To address this, we introduce a parallel Russian annotation for the large and diverse English GUM RST corpus. Leveraging recent advances, our end-to-end RST parser achieves state-of-the-art results on both English and Russian corpora. It demonstrates effectiveness in both monolingual and bilingual settings, successfully transferring even with limited second-language annotation. To the best of our knowledge, this work is the first to evaluate the potential of cross-lingual end-to-end RST parsing on a manually annotated parallel corpus.

\end{abstract}

\section{Introduction}

Discourse parsing aims to reveal the higher-level organization of text. While the task has gained significant traction in recent years, cross-lingual rhetorical structure parsing remains a complex challenge. This stems from the inherent diversity of annotation schemes across languages within the Rhetorical Structure Theory (RST) framework and the scarcity of parallel corpora. Existing large RST corpora are inconsistent in annotation guidelines, genre representation, source selection, and relation definitions. Therefore, current studies might underestimate the true potential of RST parsers for language transfer. This study addresses these challenges by introducing a Russian version of the RST part of the Georgetown University Multilayer (GUM) corpus \cite{zeldes2017gum}, encompassing all 213 original documents. This large parallel corpus provides a valuable resource for bilingual discourse analysis, enabling the development of robust RST models that can effectively capture the rhetorical structure of text in both languages.

As previous research suggests \cite{da2010comparing,iruskieta2015qualitative,cao-etal-2018-rst,cao-2020-discourse}, differences in rhetorical structures across languages primarily arise at the lower structural levels, while the global document organization exhibits some universality. Currently, top-down, unified-model frameworks \cite{nguyen-etal-2021-rst,liu-etal-2021-dmrst} have proven highly effective for end-to-end RST parsing. Hypothetically, these parsers should begin by constructing a language-independent high-level structure, with language-specific nuances incorporated primarily at lower levels.  
This study investigates the effectiveness of an end-to-end top-down RST parser adaptation across genres in a second language, utilizing both monolingual and bilingual training data. Recognizing the substantial cost of RST annotation, we further investigate the efficient amount of second-language annotation for parser transfer.

The main contributions of this work are:
\begin{enumerate}
    \item A parallel Russian annotation of a large and diverse English GUM RST corpus dubbed RRG, enabling the development and evaluation of cross-lingual RST models. This resource enables the development and evaluation of cross-lingual RST models following the same annotation framework, addressing a critical gap in the field. 
    \item A unified end-to-end RST parser achieving state-of-the-art performance on diverse benchmarks in both English and Russian:
        \begin{itemize}
            \item English: RST-DT (53.0\% end-to-end Full F1), GUM$_{9.1}$ (47.9\% F1 -- En, 47.6\% F1 -- bilingual),
            \item Russian: RRT (45.3\% F1), new RRG (44.6\% F1 -- Ru, 45.4\% F1 -- bilingual).
        \end{itemize}
\end{enumerate}

Data, code, and models are publicly available at \url{https://github.com/tchewik/BilingualRSP}. 

\section{Related Work}

Our work intersects with two key areas of RST parsing: end-to-end and cross-lingual approaches. We review prior research in this section.

\paragraph {Top-down Document-level RST Parsing}

The paradigm of top-down rhetorical parsing has recently emerged and is receiving significant attention for its exceptional capabilities for efficient end-to-end analysis through a unified model.
\citet{zhang-etal-2020-top} proposed a top-down strategy for parsing rhetorical structure from a sequence of elementary discourse units (EDUs). %Each EDU is encoded as a composite of a BiGRU hidden state and a weighted sum of token representations. Convolutional layers are employed to compute representations of potential split points between adjacent EDUs. An encoder-decoder module with an internal stack is then utilized to iteratively rank these split points, ultimately assigning each EDU to its corresponding rhetorical role. 
An encoder-decoder module with an internal stack iteratively ranks the split points, ultimately assigning each EDU to its corresponding rhetorical role.
To account for the variation in document structure context at different levels of granularity, \citet{kobayashi2020top} presented a multi-level tree construction approach developing distinct paragraph- and sentence-level discourse unit representations. Multiple monolingual language models were tested in this framework by \citet{kobayashi-etal-2022-simple}. \citet{koto-etal-2021-top} simplified the parsing by reformulating it as a sequence labeling for sequences of EDUs. 
\citet{zhang-etal-2021-adversarial} proposed computing an additional loss based on the dissimilarity between 3D representations of both gold and predicted trees, guiding the latter towards closer alignment with the original structures. Addressing the limitations of previous methods, \citet{nguyen-etal-2021-rst} devised an end-to-end document-level parsing model. This architecture presents two key advantages: (1) it seamlessly integrates tree construction and EDU segmentation through token-level splitting decisions, and (2) it employs beam search for non-greedy RST parsing. \citet{liu-etal-2021-dmrst} introduced a joint model where a shared LM encoder is employed for both segmentation and tree construction. The tree is built via attention over the sequence of EDUs within the current unit. We adopt this approach, with further details provided in Section \ref{sec:end-to-end-parser}.

\paragraph {Cross-lingual Rhetorical Parsing} 
The qualitative comparison conducted by \citet{iruskieta2015qualitative} laid the foundation for multilingual rhetorical structure analysis. Applied to a small parallel corpus across English, Spanish, and Basque (318 EDUs per language), their method revealed significant similarities in rhetorical structures between languages. Differences primarily manifested in segmentation (sentence-level discourse structure). This insight inspired subsequent efforts to bridge the gap between languages. \citet{cao-etal-2018-rst} developed a Spanish-Chinese bilingual RST Treebank consisting of 50 texts per language with varying lengths (111-1774 words). \citet{cao-2020-discourse} conducted a comparative analysis of Spanish and Chinese, identifying discourse marker and punctuation changes, EDU order variations, and EDU insertions as key contributors to sentence-level differences.
\citet{braud-etal-2017-cross} laid the groundwork for cross-lingual parsing experiments by harmonizing RST treebanks across languages and introducing 18 unified coarse-grained rhetorical labels. Subsequent work by \citet{iruskieta-braud-2019-eusdisparser} leveraged multilingual word embeddings to adapt mono- and multilingual parsers to the Basque with limited RST annotations. \citet{liu-etal-2020-multilingual-neural,liu-etal-2021-dmrst} then developed a novel neural parser utilizing EDU-level machine translation (MT). These advancements, while addressing data sparsity, also reveal challenges like ensuring the rhetorical naturalness of the texts translated segment-by-segment. The recent Georgetown Chinese Discourse Treebank (GCDT) \cite{peng-etal-2022-gcdt} offers RST annotations for 50 Chinese texts (9710 EDUs) spanning 5 of 10 genres found in the GUM corpus following the same relation inventory. Notably, 19 documents drawn from multilingual sources like Wikipedia, Wikinews, and wikiHow have English counterparts in GUM, although content and presentation may diverge across languages. 

\section{RST Corpora}
\label{sec:corpora}

\begin{table*}[ht!]
\centering
\footnotesize
\begin{tabular}{@{}llllllllllll@{}}
\toprule
\multirow{2}{*}{} & \multirow{2}{*}{Genres} & \multirow{2}{*}{Sources} & \multirow{2}{*}{Docs} & \multirow{2}{*}{Classes} & \multicolumn{3}{l}{Tokens per tree} & \multirow{2}{*}{\begin{tabular}[c]{@{}l@{}}Spanned \\ non-EDU\\ sent., \%\end{tabular}} & \multirow{2}{*}{EDUs} & \multirow{2}{*}{\begin{tabular}[c]{@{}l@{}}EDUs \\ per \\ tree\end{tabular}} & \multirow{2}{*}{\begin{tabular}[c]{@{}l@{}}Relation\\ pairs\end{tabular}} \\ \\ \cmidrule(lr){6-8}
                  &      &      &        &       & min   & max     & median &         &          &         &          \\ \midrule
RST-DT (En)       &  1   &  1   & 385    & 41    &  30   & 2624    & 396    & 79.4    & 21789    &  56.6   & 21404    \\
GUM (En)          & 12   & 12+  & 213    & 27    & 167   & 1879    & 989    & 72.5    & 26319    & 123.6   & 26106    \\
RRT (Ru)          &  2   & 17+  & 233    & 24    &   2   & 1148    &  89    & 76.7    & 28372    &  11.7   & 25957    \\
RRG (Ru)          & 12   & 12+  & 213    & 27    & 137   & 1629    & 833    & 76.9    & 25223    & 118.4   & 25010    \\ 
\bottomrule
\end{tabular}
\caption{Statistics of the corpora.}
\label{tab:corpora_stats}   
\end{table*}

This work employs three previous RST datasets for two languages: English RST-DT\footnote{\url{https://catalog.ldc.upenn.edu/LDC2002T07}; under an LDC license.} \cite{carlson-etal-2001-building} and GUM$_{9.1}$\footnote{\url{https://github.com/amir-zeldes/gum/releases/tag/V9.1.0}; CC BY 4.0.} \citep{zeldes2017gum}, Russian RuRSTreebank$_{2.1}$ \citep{pisarevskaya2017towards}. Furthermore, we suggest an additional parallel annotation for the Georgetown RST annotations (GUM$_{9.1}$) in Russian. This section discusses the datasets and preprocessing steps.

The general corpora analysis outlined in Table~\ref{tab:corpora_stats} reveals differences between the corpora extending beyond variation in genres, tree sizes, and relation labels inventory. For instance, in the RST-DT corpus, 79.4\% of non-elementary sentences\footnote{Sentence boundary prediction was performed using \texttt{spaCy} (English) and \texttt{razdel} (Russian) libraries for consistency. This approach minimizes the impact of potential errors from the sentence splitters on the comparison of the datasets. } (those containing at least one relation) are spanned by well-formed rhetorical subtrees. This high prevalence, along with explicit sentence and paragraph boundary annotation, fostered research on sentence-level RST analysis \cite{soricut-marcu-2003-sentence,joty-etal-2012-novel,nejat-etal-2017-exploring,lin-etal-2019-unified,zhang-etal-2021-language}. GUM exhibits less frequent alignment between formal sentence boundaries and rhetorical subtrees. Moreover, GUM's RST annotations used for parser training and evaluation exclude paragraph markers altogether, contrasting with the explicit boundaries present in RST-DT. %In contrast, the GUM corpus takes a different approach by ignoring formal sentence and paragraph boundaries and omitting paragraph markers altogether. 
These differences underscore that variations in the rhetorical structure, even within the same genre,\footnote{See Appendix \ref{sec:genre_edu_coverage} for genre-wise comparison.} stem not only from diverse relation sets and text sources, as \citet{liu-zeldes-2023-cant} suggest, but also from differences in annotation principles.

\subsection{Annotations for English}

\paragraph{RST-DT} 
The RST-DT corpus remains the primary benchmark for RST parsing, offering fine-grained annotations for WSJ news articles of various lengths. 

\paragraph{GUM}
The Georgetown University Multilayer corpus is an expending multi-genre corpus containing multiple layers of linguistic annotation, including RST. Featuring both written and spoken language across 12 genres, it remains the largest monolingual RST annotation corpus to date.

\subsection{RRT (RuRSTreebank)}
\label{ssec:rrt}

We exclude the scientific portion of the RuRSTreebank corpus in our experiments, as these are reported to be the first attempts at RST annotation for Russian following the earliest incompatible guidelines \cite{chistova2021rst}. The resulting dataset comprises news articles and blogs from diverse sources. It includes 5 news sources and 17 blogs covering topics such as travel, life stories, IT, cosmetics, health, politics, environment, and psychology. Despite the diversity, most documents are only partially annotated. Among the 233 document annotations, only one text is fully covered by a single tree; the remaining documents have random under-annotations. The maximum number of trees in a single \texttt{*.rs3} document reaches 42, with an average of 11.7 trees per document. This has influenced previous attempts to build a Russian parser \cite{chistova2021rst,chistova2022dialogue}, in which many efforts are directed towards predicting a look-alike forest for each full document. However, we emphasize the clear randomness of tree boundaries within the text, treating each connected tree as a separate document in our study.\footnote{
The original train/dev/test corpus splitting is preserved. The documents are only split into \texttt{docname\_part\_*.rs3} files processed independently. Documents containing only a single EDU are excluded. Within the refined corpus used for experiments, 12.8\% of trees are constructed of 2 to 4 elementary discourse units.
} Our approach’s validity is implicitly supported by the absence of rhetorical relations for higher-level textual organization (such as \textsc{Heading} or \textsc{Topic-Change}) in the RRT. Additionally, we’ve observed that in corpora for other languages, the fully annotated tree often represents only a portion of the original text. 
Following established practices in end-to-end discourse parsing for RRT, we address inconsistencies in the assignment of specific relations documented by \citet{pisarevskaya2017towards}. The dictionary in Appendix \ref{sec:rrt_relations_mapper} assists in remapping these relations during corpus preprocessing.

\subsection{RRG} 
\label{sec:rrg}

The Russian RST dataset from Georgetown University Multilayer corpus (RRG) was constructed by manually translating the RST annotations in GUM$_{9.1}$.\footnote{The train/dev/test splitting employed in the GUM corpus is preserved.} A single document required an investment of up to 2.5 hours, with the overall process consisting of:

\paragraph {Translation} We prioritized manual translation for 213 English texts, ensuring literary accuracy and genre-specific adaptation. This approach differs from the common practice in cross-lingual RST research, which often relies on EDU-level machine translation. Additionally, we ensured the precise translation of established terms and references through thorough research. Furthermore, speaker gender was identified by examining audio recordings for the \texttt{vlog} and \texttt{conversation} parts of the corpus.

\begin{figure}[h!]
\centering
    \begin{subfigure}[b]{0.27\textwidth}
        \centering
        \includegraphics[width=0.8\textwidth]{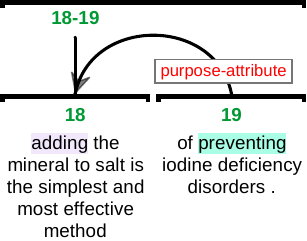}
        \par\bigskip
        \par
        \caption{GUM annotation.}
        \label{fig:rst_ex_ru_full}
    \end{subfigure}\hfill
    \begin{subfigure}[b]{0.2\textwidth}
        \centering
        \includegraphics[width=0.52\textwidth]{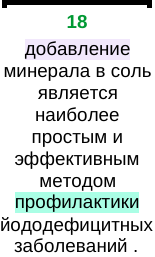}
        \caption{Corresponding RRG annotation.}
        \label{fig:rst_ex_en2ru_full}
    \end{subfigure}
    
    \caption{N-to-One EDU mapping;~\texttt{news\_iodine}.}
    \label{fig:n-to-one-mapping}
\end{figure}

\paragraph {Rhetorical Structure Alignment} The translated texts were manually aligned to the original structures unit-by-unit, following the guidelines for EDU segmentation in Russian developed for RRT.\footnote{\url{https://rstreebank.ru/eng}} To ensure consistency, an expert adjusted the annotations considering translation nuances. We added or removed elementary discourse units from the tree based on the discourse segmentation in the Russian sentences. Rhetorical relations and nuclearity were assigned following the GUM RST annotation guidelines.\footnote{\url{https://wiki.gucorpling.org/gum/guidelines}} Since our approach involved refining sentence-level relations rather than constructing trees from scratch, we anticipated minimal deviation in the annotation of rhetorical structure. As shown in Table \ref{tab:corpora_stats}, RRG contains ~95.9\% of the EDUs found in GUM.  Analysis revealed a predominant pattern of N-to-One mappings between unaligned EDUs, primarily due to language-specific differences:
\begin{itemize}
    \item English verbs sometimes translate naturally to Russian nouns, e.g., \texttt{adding} becomes \textit{\textcyrillic{добавление}}, and \texttt{preventing} translates to \textcyrillic{профилактика} (Figure \ref{fig:n-to-one-mapping}).
    \item Russian often favors active voice, collapsing passive constructions (e.g., "[there's sufficient iodine] [added into the food supply]" becomes "\textcyrillic{в пищевые продукты поступает достаточное количество йода}").
    \item Some RRG EDUs correspond to the reduced \textsc{same-unit} relations in GUM (see Appendix~\ref{sec:construction_example} for an illustration).
\end{itemize}

\begin{figure}[h!]
\centering
    \begin{subfigure}[b]{0.2\textwidth}
        \centering
        \includegraphics[width=0.93\textwidth]{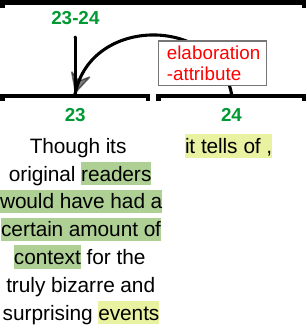}
        \par\bigskip
        \caption{GUM annotation.}
        \label{fig:rst_ex_ru_full}
    \end{subfigure}\hfill
    \begin{subfigure}[b]{0.28\textwidth}
        \centering
        \includegraphics[width=\textwidth]{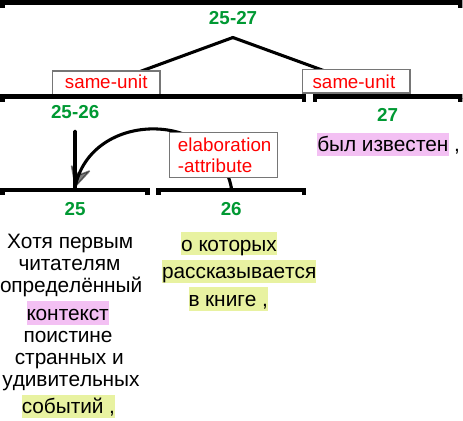}
        \caption{Corresponding RRG annotation.}
        \label{fig:rst_ex_en2ru_full}
    \end{subfigure}
    
    \caption{One-to-N EDU mapping;~\texttt{fiction\_wedding}.}
    \label{fig:one-to-n-mapping}
\end{figure}

One-to-N mappings occur when a single EDU in GUM is split into multiple units in RRG. This is primarily observed with prepositional phrases and instances where the Russian syntax allows for greater variation (Figure \ref{fig:one-to-n-mapping}).

\paragraph {Annotation Polishing} Our efforts to detect and correct misassigned labels and misaligned EDUs in the RRG draft began with an examination of the class distribution. It helped us identify obvious annotation errors, including some inherited from the original English corpus (such as rare and unlikely classes like \textsc{Restatement\_SN}). To further refine the annotations, we trained the RST label classifier for Russian proposed by \citet{chistova2021rst} on the draft dataset. This classifier served as an outlier detection tool, allowing us to detect potentially mislabeled examples. Specifically, we focused on cases where the classifier confidently predicted an incorrect class and excluded the true (annotated) class from its top 3 most probable predictions. Following the GUM relation annotation guidelines, we fixed any corrupted structures identified through this analysis. This process also revealed minor inherited annotation inconsistencies, which we standardized in the final RRG dataset (see Figure \ref{tab:standardization_gum} Appendix \ref{sec:rurstg_details} for details). 

% \section{Preamble}
% \end{quote}

\section{End-to-End RST Parser}
\label{sec:end-to-end-parser}

The rhetorical structure parsers suggested in recent years \cite{zhang-etal-2020-top,kobayashi2020top,zhang-etal-2021-adversarial,nguyen-etal-2021-rst} often focused on developing innovative features to address either specific aspects of the structure construction or its global optimization. However, these approaches often overlook the integration of previously established effective features. They also frequently neglect the end-to-end performance, a fundamental aspect of any practical framework. We are building a hybrid deep model solving both segmentation and tree construction that benefits from the techniques suggested by recent work.

\subsection{Base Model}

\begin{figure}[h!]
\centering
    \includegraphics[width=0.35 \textwidth]{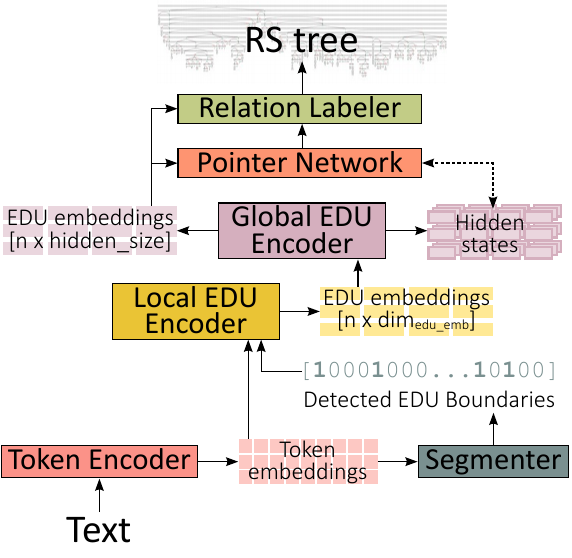}
    \caption{Architectural overview of DMRST.}
    \label{fig:dmrst}
\end{figure}

As a base end-to-end deep model, we use the DMRST \cite{liu-etal-2021-dmrst} architecture visualized in Figure \ref{fig:dmrst}. 

The framework consists of four main modules: (1) EDU segmentation via document-level labeling, (2) hierarchical EDU encoding, (3) span-splitting decoding for tree construction, and (4) nuclearity-relation prediction using a bi-affine classifier. The encoded EDU sequence is iteratively parsed during decoding, and the classifier predicts the nuclearity and relations between adjacent units. Training minimizes the dynamic weighted average (DWA) \cite{liu2019end} of losses for EDU segmentation, tree structure parsing, and nuclearity+relation labeling.

\subsection{Modifications to the Base Model}

To improve end-to-end parsing performance, we introduce modifications to the base model, focusing primarily on EDU segmentation and encoding.

\paragraph{Segmentation: \texttt{ToNy}}
 The BiLSTM-CRF segmenter known by this name \cite{muller-etal-2019-tony} is a simple yet robust neural token labeler that took first place in the DISRPT 2019 shared task \cite{zeldes-etal-2019-disrpt}. The original DMRST parser implements a feedforward token classifier (with an additional similar classifier for the right neighbor only for loss penalization).\footnote{Directly comparing segmentation scores from the report with ToNy's paper raises concerns due to differing methodological choices. DMRST employs a different pretrained language model, potentially augmented data, and document-level segmentation, contrasting with ToNy's reliance on the StanfordNLP sentence splitter. Furthermore, the original ToNy functions as a standalone segmenter, while DMRST incorporates segmentation into its unified encoder training for joint optimization with tree construction.} We replace the original DMRST segmentation module with a BiLSTM-CRF layer without additional losses.

\paragraph{Local EDU Encoding: \texttt{E-BiLSTM}}
Rather than averaging subword embeddings for local EDU encoding as in the original method, we utilize another BiLSTM layer, which enables us to achieve better sequence encodings. The concatenation of hidden states at the final time step of each pass captures the context of the phrase more precisely than an average of its subword embeddings.

\paragraph{No augmentations} One of the distinctive features of the original D\underline{M}RST is data augmentation using corpora unification and EDU-level machine translation. However, we emphasize that annotated corpora for different languages can present different interpretations of RST with nuances in the tree constraints and relation definitions. Furthermore, EDU-level MT can result in unnatural discourse structures in the target language and offer little linguistic knowledge (although it can augment examples of some relations in the training set). Therefore, we do not consider either corpora unification or machine translation. Instead, we build a full parallel RST corpus with consistent relation inventory.

\paragraph{DWA Window Size}
Dynamic weighting is crucial for ensuring that each component of the parser receives the necessary attention during training:

\begin{equation} \label{eq:2}
\mathcal{L}_{total} = \sum_{k=1}^{3} {\lambda_k \mathcal{L}_k}, \textrm{ }  w_k(i - 1) = \frac{\mathcal{L}_k(i-1)}{\mathcal{L}_k(i-2)}
\end{equation}

\begin{equation} \label{eq:3}
\lambda_k(i) = \textrm{softmax}(\frac{w_k(i-1)}{Temp}) \times 3 ,
\end{equation}
where the loss $\mathcal{L}_{total}$ is the DWA of task-specific losses with weights $\lambda_i$; $w_k$ are the relative descending rates for tasks 1 (segmentation), 2 (tree construction), and 3 (relation labeling), $i$ is an iteration index, and $Temp$ controls the softness of the task weighting.
However, relying solely on the last two batches (Equation \ref{eq:2}) is susceptible to local trend amplification, especially with smaller batches encompassing rhetorical trees of varying sizes and complexities. To address this issue, we introduce a DWA window size parameter $b$:

\begin{equation}
w_k(i - 1) = \frac{\sum_{j=1}^{b}{\mathcal{L}_k(i-j)}}{\sum_{j=b+1}^{2b}{\mathcal{L}_k(i-j)}}
\end{equation}
By analyzing a broader range of loss values, the model can effectively identify long-term trends and adjust task weights accordingly. This modification improved training stability with smaller batches, particularly on the RRT dataset comprising a large number of single-relation discourse trees.

\begin{table*}[ht]
\centering
\scriptsize
\begin{tabular}{@{}lllllll@{}}
\toprule
                     & Corpus                   & Method                                                      & S                   & N                   & R                   & Full                \\ \midrule
\multirow{21}{*}{En} & \multirow{19}{*}{RST-DT} & Human                                                       & 78.7                & 66.8                & 57.1                & 55.0                \\ 
                     \cmidrule(l){3-7}
                     &                          & \citet{feng-hirst-2014-linear}                              & 68.6                & 55.9                & 45.8                & 44.6                \\
                     &                          & DPLP \citeyearpar{ji-eisenstein-2014-representation}        & 64.1                & 54.2                & 46.8                & 46.3                \\
                     &                          & CODRA \citeyearpar{joty-etal-2015-codra}                    & 65.1                & 55.5                & 45.1                & 44.3                \\
                     &                          & \citet{surdeanu-etal-2015-two}                              & 65.3                & 54.2                & 45.1                & 44.2                \\
                     &                          & \citet{li-etal-2016-discourse}                              & 64.5                & 54.0                & 38.1                & 36.6                \\
                     &                          & HILDA \citeyearpar{hayashi-etal-2016-empirical}             & 65.1                & 54.6                & 44.7                & 44.1                \\
                     &                          & \citet{braud-etal-2016-multi}                               & 59.5                & 47.2                & 34.7                & 34.3                \\
                     &                          & \citet{braud-etal-2017-cross}                               & 62.7                & 54.5                & 45.5                & 45.1                \\
                     &                          & \citet{yu-etal-2018-transition}                             & 71.4                & 60.3                & 49.2                & 48.1                \\
                     &                          & \citet{mabona-etal-2019-neural}                             & 67.1                & 57.4                & 45.5                & 45.0                \\
                     &                          & \citet{zhang-etal-2020-top}                                 & 67.2                & 55.5                & 45.3                & 44.3                \\
                     &                          & \citet{nguyen-etal-2021-rst}                                & 74.3                & 64.3                & 51.6                & 50.2                \\
                     &                          & \citet{koto-etal-2021-top}                                  & 73.1                & 62.3                & 51.5                & 50.3                \\
                     &                          & \citet{zhang-etal-2021-adversarial}                         & 76.3                & 65.5                & 55.6                & 53.8                \\
                     &                          & DMRST + Cross-translation \citeyearpar{liu-etal-2021-dmrst} & 76.7                & 66.2                & 56.5                & --                  \\
                     &                          & \citet{yu-etal-2022-rst}                                    & 76.4                & 66.1                & 54.5                & 53.5                \\
                     &                          & \citet{kobayashi-etal-2022-simple}                          & 77.8 ± 0.3          & \textbf{68.0 ± 0.5} & \textbf{57.3 ± 0.2} & 55.4 ± 0.4          \\
                     \cmidrule(l){3-7} 
                     &                          & DMRST (this work)                                           & \textbf{78.7 ± 0.4} & \textbf{68.0 ± 0.6} & \textbf{57.3 ± 0.2} & \textbf{55.7 ± 0.3} \\ 
                     &                          & + \texttt{ToNy}                                             & 78.4 ± 0.7          & 67.4 ± 0.8          & 56.8 ± 0.9          & 55.2 ± 0.9          \\ 
                     &                          & + \texttt{ToNy} + \texttt{E-BiLSTM}                         & 78.5 ± 0.5          & 67.5 ± 0.7          & 57.0 ± 0.5          & 55.3 ± 0.5          \\ 
    \cmidrule(l){2-7} 
                     & \multirow{2}{*}{GUM v9.1} & DMRST (this work)                                          & 72.7 ± 0.7          & 60.8 ± 0.6          & 52.8 ± 0.5          & 51.7 ± 0.4          \\
                     &                          & + \texttt{ToNy}                                             & 72.8 ± 0.3          & \textbf{61.4 ± 0.6} & \textbf{53.1 ± 0.5} & \textbf{52.0 ± 0.5} \\ 
                     &                          & + \texttt{ToNy} + \texttt{E-BiLSTM}                         & \textbf{73.1 ± 0.3} & 61.3 ± 0.2          & 53.0 ± 0.3          & \textbf{52.0 ± 0.3} \\
\midrule
\multirow{4}{*}{Ru}  & \multirow{2}{*}{RRT}     & DMRST (this work)                                           & 81.0 ± 0.5          & 63.3 ± 0.9          & 54.2 ± 0.9          & 54.0 ± 0.9          \\
                     &                          & + \texttt{ToNy}                                             & 80.9 ± 1.0          & \textbf{63.4 ± 0.9} & \textbf{54.7 ± 0.9} & \textbf{54.6 ± 0.9} \\ 
                     &                          & + \texttt{ToNy} + \texttt{E-BiLSTM}                         & \textbf{81.2 ± 0.4} & 62.9 ± 0.9          & 53.8 ± 1.2          & 53.6 ± 1.2          \\ 
    \cmidrule(l){2-7} 
                     & \multirow{2}{*}{RRG}     & DMRST (this work)                                           & \textbf{71.5 ± 0.4} & \textbf{57.6 ± 0.2} & \textbf{49.1 ± 0.3} & \textbf{47.9 ± 0.2} \\
                     &                          & + \texttt{ToNy}                                             & 71.1 ± 0.5          & 56.6 ± 1.4          & 48.2 ± 1.5          & 47.2 ± 1.4          \\ 
                     &                          & + \texttt{ToNy} + \texttt{E-BiLSTM}                         & 70.7 ± 0.4          & 56.4 ± 0.5          & 48.3 ± 0.5          & 47.1 ± 0.5          \\ 
\bottomrule
\end{tabular}
\caption{RST parsing performance evaluated on the gold EDU segmentation. Micro F1 scores (original Parseval); average and standard deviation. Missing values are not reported in the cited work.}
\label{tab:all_gold}
\end{table*}  

\begin{table*}[]
\centering
\scriptsize
\setlength{\tabcolsep}{3pt}
\begin{tabular}{lllccccc}
\toprule
                    & Corpus                   & Method  & Segm.               & S                   & N                   & R                   & Full                \\ \midrule
\multirow{8}{*}{En} & \multirow{5}{*}{RST-DT}  & SegBot \citeyearpar{li2018segbot} \& \citet{zhang-etal-2020-top} & 92.2 & 62.3 & 50.1 & 40.7 & 39.6 \\
                    &                          & \citet{nguyen-etal-2021-rst}             & 96.3        & 68.4        & 59.1        & 47.8        & 46.6                \\ 
                    &                          & DMRST \citeyearpar{liu-etal-2021-dmrst}  & 96.4        & 69.8        & 59.4        & 49.4        & 48.6                \\
                    &                          & + Cross-translation                      & 96.5        & 70.4        & 60.6        & 51.6        & 50.1                \\  
                    \cmidrule(l){3-8}
                    &                          & DMRST (this work)                        & 97.3 ± 0.1          & 74.3 ± 0.6          & 64.1 ± 0.7          & 53.9 ± 0.5          & 52.4 ± 0.5          \\
                    &                          & + \texttt{ToNy}                          & \textbf{97.9 ± 0.1} & \textbf{75.1 ± 0.7} & \textbf{64.8 ± 0.7} & \textbf{54.5 ± 0.9} & \textbf{53.0 ± 0.9} \\ 
                    &                          & + \texttt{ToNy} + \texttt{E-BiLSTM}      & 97.8 ± 0.1          & 74.8 ± 0.5          & 64.5 ± 0.8          & \textbf{54.5 ± 0.7} & \textbf{53.0 ± 0.7} \\
\cmidrule{2-8} 
                    & \multirow{2}{*}{GUM v9.1}& DMRST (this work)                        & 94.7 ± 0.4          & 65.0 ± 0.5          & 54.2 ± 0.5          & 47.3 ± 0.5          & 46.4 ± 0.4          \\
                    &                          & + \texttt{ToNy}                          & 95.4 ± 0.1          & 66.4 ± 0.3          & 55.8 ± 0.5          & 48.5 ± 0.5          & 47.6 ± 0.6          \\ 
                    &                          & + \texttt{ToNy} + \texttt{E-BiLSTM}      & \textbf{95.5 ± 0.1} & \textbf{66.9 ± 0.5} & \textbf{56.1 ± 0.3} & \textbf{48.8 ± 0.4} & \textbf{47.9 ± 0.4} \\                     \midrule
\multirow{5}{*}{Ru} & \multirow{2}{*}{RRT}     & DMRST (this work)                        & \textbf{92.4 ± 0.3} & \textbf{66.5 ± 1.0} & \textbf{52.4 ± 1.2} & \textbf{45.3 ± 1.0} & \textbf{45.3 ± 1.0} \\
                    &                          & + \texttt{ToNy}                          & \textbf{92.4 ± 0.2} & 65.4 ± 1.1          & 51.3 ± 0.6          & 44.6 ± 0.5          & 44.5 ± 0.5          \\
                    &                          & + \texttt{ToNy} + \texttt{E-BiLSTM}      & 92.2 ± 0.2 & 65.9 ± 0.5          & 51.0 ± 0.7          & 43.9 ± 1.0          & 43.8 ± 1.0          \\ \cmidrule{2-8} 
                    & \multirow{2}{*}{RRG}     & DMRST (this work)                        & 96.3 ± 0.1          & 65.6 ± 0.3          & 52.8 ± 0.3          & 45.1 ± 0.2          & 44.0 ± 0.3          \\
                    &                          & + \texttt{ToNy}                          & 96.7 ± 0.2          & \textbf{66.6 ± 0.9} & 53.0 ± 1.7          & 45.3 ± 1.7          & 44.3 ± 1.5          \\
                    &                          & + \texttt{ToNy} + \texttt{E-BiLSTM}      & \textbf{96.9 ± 0.2}  & 66.5 ± 0.4         & \textbf{53.3 ± 0.6} & \textbf{45.8 ± 0.5} & \textbf{44.6 ± 0.4} \\
\bottomrule
\end{tabular}
\caption{End-to-end parsing performance. Micro F1 scores (original Parseval); average and standard deviation.}
\label{tab:all_end2end}
\end{table*}

\section{Experimental Setup}

In this study, we adopt the multilingual \texttt{xlm-roberta-large}\footnote{MIT License.} \cite{conneau-etal-2020-unsupervised} model known for its exceptional zero-shot performance across discourse relation classification tasks in multiple languages \cite{kurfali-ostling-2021-probing}. Hyperparameters are fixed as specified in Appendix~\ref{sec:implementation}. We average results across five runs with varying model seeds (fixed-split corpora: GUM and RRG, RRT) or different train/dev splits (RST-DT). Bilingual experiments (Section~\ref{sec:cross-lingual}) additionally involve randomly selecting 25\%, 50\%, and 75\% of the second-language data for each of the five runs.

\section{Monolingual Evaluation and Discussion}
\label{sec:models_evaluation}

This section evaluates the monolingual parsing performance for two languages. Our baseline DMRST (this work) differs from the DMRST \citeyearpar{liu-etal-2021-dmrst} by employing the \texttt{xlm-roberta-large} language model and DWA window size parameter.

\subsection {Segmentation}

Segmentation performance is shown in Table~\ref{tab:all_end2end} alongside other metrics for end-to-end parsing.

\paragraph{English}
The previous best segmentation performance belongs to the DisCut\footnote{A simple token classifier for sentences on top of the \texttt{XLM-RoBERTa-large}.} method \cite{metheniti-etal-2023-discut}, achieving 97.6\% F1 on RST-DT\footnote{
Inter-annotator agreement for segmentation on a subset of 53 \cite{carlson-etal-2001-building} double-annotated texts within the RST-DT corpus yielded a score of 98.3\% F1 \cite{soricut-marcu-2003-sentence}. However, this evaluation remains limited to a small part of the corpus that does not align with its test section. The human agreement scores reported in Table \ref{tab:all_gold} are obtained on the same part of the corpus \citep{joty-etal-2015-codra}.
} and 95.5\% F1 on GUM$_{9.0}$. 
Our improved DMRST\texttt{+ToNy} surpasses this on RST-DT with an average of 97.9\% F1. The final model also outperforms our baseline on GUM$_{9.1}$ reaching an average F1 score of 95.5\% compared to 94.7\%.

\paragraph{Russian}
Building upon the ToNy \citeyearpar{muller-etal-2019-tony} method, \citet{chistova2022dialogue} achieve an F1 score of 89.1\% on the RRT$_{2.1}$ corpus%\footnote{Their original paper reports an F1 score of 88.4\% for EDU segmentation on the RuRSTreebank corpus (v2.0) using the then-latest version of the parser. The current results pertain to the latest version of both the corpus (v2.1) and the parser.}
). The DISRPT shared tasks \citeyearpar{zeldes-etal-2019-disrpt,zeldes-etal-2021-disrpt,braud-etal-2023-disrpt} featured an early and flawed version of RRT, which had non-hierarchical annotations of academic genres. Thus, the performance in segmentation and relation classification reported for their version of the dataset is not consistent with the version used in the current work on end-to-end discourse parsing for Russian. The details on the current version (RRT$_{2.1}$) are outlined in Section \ref{ssec:rrt}.
While the architecture modifications did not significantly impact segmentation performance on the RRT, they consistently improved it on the RRG corpus, with an average increase from 96.3\% F1 to 96.9\% F1.

\subsection {Assessing the Joint Model}

Our experiment on joint training of segmentation and parsing modules within a unified architecture produced intriguing results, revealing a fundamental tension between the two tasks. 
Models with higher F1 scores on gold-standard segmentation (Table \ref{tab:all_gold})  performed worse on both segmentation and end-to-end parsing metrics than models with lower gold-segmentation scores but better utilization of their predicted segments (Table \ref{tab:all_end2end}). 
This pattern suggests that the encoder representations are being pulled in two opposing directions during fine-tuning.  Sentence segmentation relies heavily on local cues within sentences, leading segmentation-optimized models to develop encodings for fine-grained syntactic patterns. However, building a document-level parse tree requires capturing long-range context and global relationships, demanding encodings that recognize complex discourse units.
Therefore, directly comparing jointly trained models on gold-EDU trees may not be reliable in this scenario. The following discussion delves into the end-to-end parsing evaluated in Table \ref{tab:all_end2end}.

\paragraph{English} The enhanced models achieve state-of-the-art results for end-to-end English RST parsing. Leveraging \texttt{ToNy} segmentation for the RST-DT dataset and both \texttt{ToNy} and BiLSTM EDU encoding for the GUM dataset, we obtain a substantial improvement in unlabeled tree construction, measured by the Span metric (average increase of 0.8\% for RST-DT and 1.9\% for GUM). This gain is noteworthy considering the widespread use of unlabeled rhetorical trees in RST parsing applications \cite{guzman-etal-2014-using,khosla-etal-2021-evaluating}. Nuclearity assignment, crucial for tasks like summarization and sentiment analysis \cite{goyal-eisenstein-2016-joint,fu2016long,huber-carenini-2020-mega}, also benefits from our approach. The best models achieve an average F1-score of 64.8\% (+0.7\%) on RST-DT and 56.1\% (+1.9\%) on GUM for the Nuclearity metric. Finally, the full rhetorical structure construction for both datasets achieves 53.0\% for RST-DT and 47.9\% for GUM. 

\paragraph{Russian} While the enhanced model noticeably improved performance on other corpora, it surprisingly failed to do so on RRT. This disparity might be attributed to the overfitting of the ToNy segmenter, potentially caused by the larger batch size necessary for stable RRT training (Appendix \ref{sec:implementation}). Fewer EDUs per tree in RRT (Table~\ref{tab:corpora_stats}) lead to shallower, less complex structures, maximizing the Span score for gold-standard segmentation (81.2\% for the best model in Table~\ref{tab:all_gold}).
Building trees from EDUs predicted with 92\% F1 (Table \ref{tab:all_end2end}) significantly drops the Span metric (15\% F1 gap). Similar to the original GUM corpus, the model incorporating both modifications achieved the best results on RRG, exhibiting an average Full end-to-end F1-score of 44.6\%.

\begin{table*}[ht]
\centering
\scriptsize
\begin{tabular}{@{}rlllllllllll@{}}
\toprule
\multicolumn{1}{l}{\multirow{2}{*}{En}} & \multicolumn{1}{l}{\multirow{2}{*}{Ru}} & \multicolumn{5}{l}{En}                                         & \multicolumn{5}{l}{Ru}                                         \\ \cmidrule(l){3-12} 
\multicolumn{1}{l}{}                    & \multicolumn{1}{l}{}                    & Segm.      & S          & N          & R          & Full       & Segm.      & S          & N          & R          & Full       \\ \midrule
\multirow{5}{*}{100\%}                  & 0\%                                     & 95.5 ± 0.1 & 66.9 ± 0.5 & 56.1 ± 0.3 & 48.8 ± 0.4 & 47.9 ± 0.4 & 95.5 ± 0.3 & 63.9 ± 0.7 & 51.4 ± 1.0 & 43.4 ± 0.6 & 42.2 ± 0.6 \\
                                        & 25\%                                    & 95.5 ± 0.1 & 66.4 ± 0.7 & 55.1 ± 1.0 & 48.2 ± 1.0 & 47.4 ± 1.0 & 96.4 ± 0.3 & 66.3 ± 0.6 & 53.8 ± 0.6 & 45.9 ± 0.7 & 44.9 ± 0.6 \\
                                        & 50\%                                    & 95.5 ± 0.1 & 66.6 ± 0.5 & 55.4 ± 0.6 & 48.7 ± 0.6 & 47.7 ± 0.7 & 96.6 ± 0.2 & 67.0 ± 0.5 & 54.2 ± 0.6 & 46.6 ± 0.8 & 45.5 ± 0.8 \\
                                        & 75\%                                    & 95.6 ± 0.2 & 67.2 ± 0.2 & 55.7 ± 0.5 & 48.9 ± 0.6 & 47.9 ± 0.5 & 96.8 ± 0.2 & 67.0 ± 0.4 & 54.0 ± 0.5 & 46.2 ± 0.5 & 45.0 ± 0.5 \\
                                        & 100\%                                   & 95.3 ± 0.1 & 66.4 ± 0.7 & 55.2 ± 0.6 & 48.6 ± 0.6 & 47.6 ± 0.7 & 96.8 ± 0.1 & 66.9 ± 0.4 & 54.3 ± 0.3 & 46.5 ± 0.4 & 45.4 ± 0.4 \\ \bottomrule
\end{tabular}
\caption{Performance of the models trained with second language data injection.}
\label{tab:transfer_results}
\end{table*}

\section{Cross-Dataset Compatibility in Russian RST Parsing}

This section explores the cross-dataset compatibility of Russian RST parsing by comparing two relation inventories derived from RRT and RRG parsers using a data-driven approach.

\paragraph{Relation Labeling}
To categorize the discourse unit pairs connected in the annotated corpora, we trained the relation classifier for Russian developed by \citet{chistova2021rst}. It is an ensemble of a feature-rich classifier and an ELMo-driven classifier. The feature-rich classifier includes a comprehensive dictionary of discourse cues in Russian, various morpho-syntactic features, a sentiment classifier, and USE vectors \cite{cer2018use}. The neural classifier is based on the BiMPM architecture \cite{wang2017bimpm}, and utilizes the ELMo model for Russian as well as pre-trained fastText embeddings \cite{bojanowski-etal-2017-enriching} and character n-gram embeddings to encode a discourse unit. 
The RRT dataset, which includes 24 classes, yielded a 48.9\% macro F1 score, while the RRG dataset, which includes 27 classes, yielded a 46.3\% macro F1 score (see Appendix \ref{sec:clf_details} for detailed results). Cross-dataset classification results illustrated in Appendix \ref{sec:clf_details} Figure \ref{fig:clf_transfer_results} indicate a notable overlap among the majority of classes from the two datasets while also highlighting the challenge of RST treebanks unification across languages and frameworks. 

\section{Cross-Lingual Evaluation}
\label{sec:cross-lingual}

In this section, we explore the capabilities of our best \texttt{+ToNy+E-BiLSTM} model in two scenarios: (1) its performance on an unseen or under-annotated language, and (2) its bilingual adaptation when trained on a fully-annotated parallel corpus. We assess the performance of a model on a new language, analyzing how expanding the parallel training data influences its ability to parse diverse writing and speech styles. With the English training data held constant, we investigate its ability to adapt to different genres in Russian.

\paragraph{Direct Transfer} By employing documents that differ only in language, we isolate the impact of language on RST parsing within zero-shot generalization, offering a more nuanced evaluation compared to typical mixed-source approaches. As demonstrated in Table \ref{tab:transfer_results}, the RST parser achieves remarkable results on Russian test documents in a zero-shot setting (0\%), showcasing the strength of multilingual language models. It performs nearly on par with the monolingual parser specifically trained on Russian data (RRG, Table \ref{tab:all_end2end}). Although the Russian parser exhibits improvements across all metrics (segmentation: +1.4\%, Span splitting: +2.6\%, Nuclearity assignment: +1.9\%, Full: +2.4\%), the gap remains relatively narrow, demonstrating the effectiveness of the original GUM-based parser across languages. 
Reversing the direction (Russian to English) revealed a substantial performance drop (Table \ref{tab:zeroshot_all}, Appendix \ref{sec:genrewise_evaluation}). Its F1 score for English segmentation is only 86.9\%. This disparity likely stems from heavy reliance on commas to separate elementary discourse units in Russian (examples in Figure \ref{fig:clf_transfer_commas}, Appendix \ref{sec:genrewise_evaluation}).  With only 18.5\% of EDUs ending with commas in GUM compared to a staggering 37.5\% in RRG, the segmenter became overly reliant on a feature less common in English.% Adding to the challenge, RRG's segmentation guideline requires verbal forms (excluding verbal nouns) in EDUs. This rule applies to 98\% of RRG EDUs but only 85\% of GUM EDUs, reflecting both inherent language differences as well as GUM's specific annotation guidelines (examples in Figure 17, Appendix D).

\paragraph{Mixed Train Data} 
The objective of this experiment is to estimate the data requirements for successful cross-lingual transfer in RST parsing, a task that relies on laborious expert annotation. We evaluate cross-lingual transfer performance across different amounts of annotation, ranging from 25\% to 100\% of the target language corpus. Our evaluation considers an ideal scenario involving full parallel data.
Table \ref{tab:transfer_results} presents the model's performance as the number of labeled examples in the second language increases. We observe a gradual improvement in the model's ability to construct rhetorical trees with attached nuclearities. However, the rhetorical labeling accuracy plateaus at approximately 50\% of second language annotations.  The genre-specific performance of the model is illustrated in Figure \ref{fig:transfer_results}. A more detailed evaluation is provided in Appendix \ref{sec:genrewise_evaluation}. Genres such as \textit{wikihow}, \textit{textbook}, \textit{academic}, \textit{voyage}, \textit{bio} (Wikipedia), \textit{speech}, \textit{interview}, and \textit{news} exhibit the highest adaptation to the second language. Spoken discourse genres achieved the lowest parsing scores but showed notable adaptation (\textit{vlog}: 33.3\% to 36.6\% F1; \textit{conversation}: 22.1\% to 27.4\% F1).    

\begin{figure}[h!]
\centering
    \includegraphics[width=0.47\textwidth]{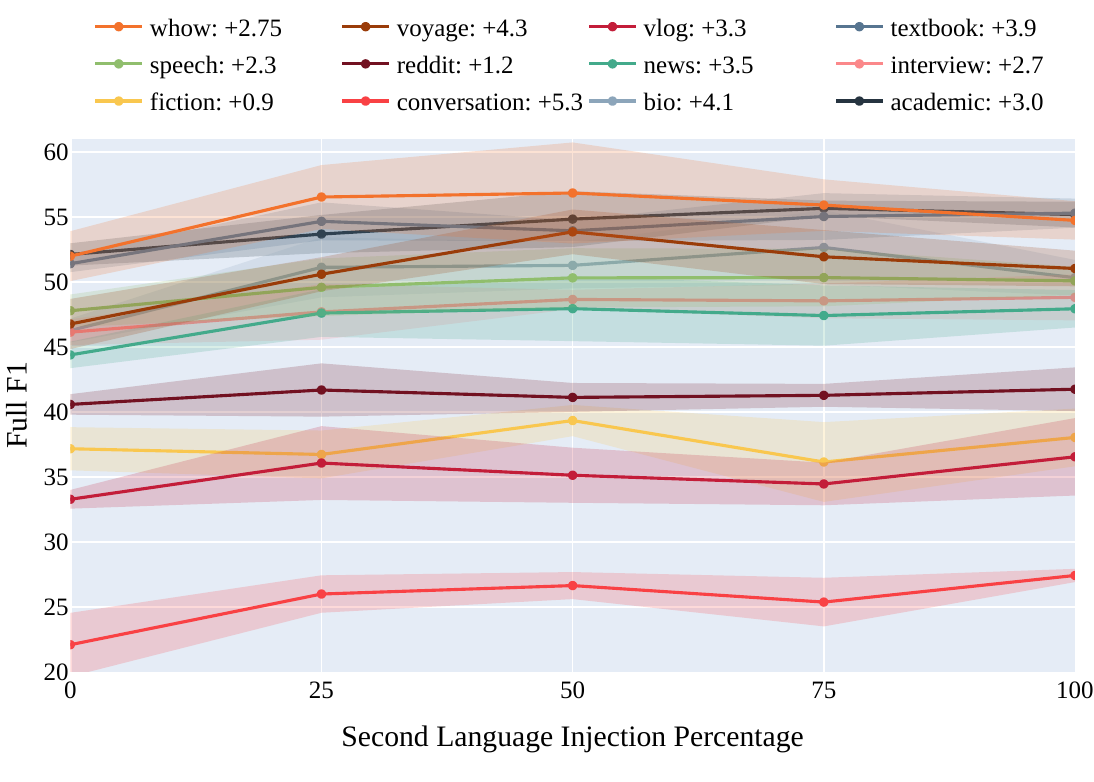}
    \caption{Impact of second language injection on the end-to-end Full performance.}
    \label{fig:transfer_results}
\end{figure}

The bilingual model outperforms the monolingual RRG model (44.6\% F1), achieving a Full end-to-end score of 45.4\% F1. This improvement might be attributed to the potential limitations of the pre-trained model, XLM-RoBERTa, in handling Russian due to the imbalanced nature of the CC-100 pre-training corpus (23.4B Russian tokens vs. 55.6B English tokens \cite{conneau-etal-2020-unsupervised}). Bilingual injection, where both languages are presented together during training, could help mitigate this imbalance by allowing the model to learn richer representations of Russian text. Despite a slight F1 decrease in English, the bilingual parser excelled in 9 out of 12 genres in Russian (as detailed in Table \ref{tab:monobilingual}), highlighting the effectiveness of bilingual training for cross-lingual transfer.

\begin{table}[]
\centering
\scriptsize
\begin{tabular}{@{}lll|lll@{}}
\toprule
Test Language & \multicolumn{2}{l|}{English}              & \multicolumn{3}{l}{Russian}               \\
Train Data    & GUM  & GUM+RRG       & GUM   & RRG  & GUM+RRG        \\ \midrule
\textit{academic}     & 56.3 & 55.5  \>({\color[HTML]{CB0000} --0.8}) & 52.1 & 55.7 & 55.2  \>({\color[HTML]{CB0000} --0.5}) \\
\textit{bio}          & 51.5 & 52.5  \>({\color[HTML]{036400} +1.0})  & 46.3 & 52.2 & 50.3  \>({\color[HTML]{CB0000} --1.9}) \\
\textit{conversation} & 29.3 & 30.2  \>({\color[HTML]{036400} +0.9})  & 22.1 & 25.9 & 27.4  \>({\color[HTML]{036400} +1.5}) \\
\textit{fiction}      & 38.5 & 40.2  \>({\color[HTML]{036400} +1.7})  & 37.2 & 36.7 & 38.0  \>({\color[HTML]{036400} +1.3}) \\
\textit{interview}    & 55.1 & 54.7  \>(--0.4)                        & 46.1 & 47.3 & 48.8  \>({\color[HTML]{036400} +1.5}) \\
\textit{news}         & 55.0 & 52.9  \>({\color[HTML]{CB0000} --2.1}) & 44.4 & 45.9 & 47.9  \>({\color[HTML]{036400} +2.0}) \\
\textit{reddit}       & 44.0 & 42.3  \>({\color[HTML]{CB0000} --1.7}) & 40.6 & 41.5 & 41.8  \>(+0.3)                        \\
\textit{speech}       & 57.6 & 57.2  \>(--0.4)                        & 47.8 & 50.2 & 50.1  \>(--0.1)                       \\
\textit{textbook}     & 57.0 & 56.4  \>({\color[HTML]{CB0000} --0.6}) & 51.4 & 53.6 & 55.3  \>({\color[HTML]{036400} +1.7}) \\
\textit{vlog}         & 41.7 & 40.6  \>({\color[HTML]{CB0000} --1.1}) & 33.3 & 35.5 & 36.6  \>({\color[HTML]{036400} +1.1}) \\
\textit{voyage}       & 44.1 & 43.4  \>({\color[HTML]{CB0000} --0.7}) & 46.8 & 49.3 & 51.0  \>({\color[HTML]{036400} +1.7}) \\
\textit{whow}         & 57.0 & 56.8  \>(--0.2)                        & 52.0 & 54.1 & 54.7  \>({\color[HTML]{036400} +0.6}) \\ 
\\
\textit{all}                   & 47.9 & 47.6  \>(--0.3)                        & 42.2 & 44.6 & 45.4  \>({\color[HTML]{036400} +0.8}) \\ \bottomrule
\end{tabular}
\caption{Mono- vs. bilingual model evaluation (avg. end-to-end Full F1).}
\label{tab:monobilingual}
\end{table}

\section{Conclusion}

This study addresses the challenges of cross-lingual discourse parsing. We introduce a large parallel Russian annotation of the multigenre GUM RST corpus and assess the performance of an end-to-end top-down model in bilingual rhetorical structure parsing. The top-down unified parser employing a multilingual language model established a strong baseline on end-to-end parsing in both languages. 
Further analysis explored direct parser transfer without second-language data. Surprisingly, transferring the English parser to Russian achieved comparable quality to the monolingual parser. However, the reverse transfer suffered due to nuances in Russian discourse segmentation, underlining the critical role of language-specific features in language transfer.
We investigated the effectiveness of porting the analyzer with limited second-language data. Our findings demonstrate that even with minimal data, such transfer remains effective. Finally, training the bilingual parser on the entire parallel dataset yielded the best discourse parsing performance in Russian, and strong performance in English.

\section*{Limitations}

While the written sections of the corpus are well-adapted to Russian, accurately capturing the nuances of Russian spontaneous speech in documents outlining English spoken discourse (\textit{vlog}, \textit{conversation}) through translation can be challenging. This presents an exciting opportunity for future research to explore the unique RST features of spoken discourse in Russian.

\bibliography{anthology,custom}
\bibliographystyle{acl_natbib}

\appendix

\section{Sentence Subtrees Coverage}
\label{sec:genre_edu_coverage}

Examining tree-covered non-elementary sentences in the analyzed corpora (see Table \ref{tab:genre_edu_coverage}) reveals evident disparities in formal structure between annotation schemas, even within the recurring \textit{news} genre.

\begin{table}[h]
\small
\centering
\begin{tabular}{@{}llcc@{}}
\toprule
Corpus & Genre        & En   & Ru   \\ \toprule
RST-DT & \textit{news}         & 79.4 & --   \\ \midrule
GUM,   & \textit{academic}     & 72.0 & 76.9 \\
RRG    & \textit{bio}          & 61.1 & 72.2 \\
       & \textit{conversation} & 65.8 & 68.7 \\
       & \textit{fiction}      & 70.4 & 78.5 \\
       & \textit{interview}    & 71.4 & 78.1 \\
       & \textit{news}         & 69.0 & 79.2 \\
       & \textit{reddit}       & 73.0 & 77.4 \\
       & \textit{speech}       & 85.8 & 87.5 \\
       & \textit{textbook}     & 78.5 & 76.4 \\
       & \textit{vlog}         & 75.3 & 77.3 \\
       & \textit{voyage}       & 71.3 & 71.5 \\
       & \textit{whow}         & 77.5 & 78.4 \\ \midrule
RRT    & \textit{blogs}        & --   & 71.6 \\
       & \textit{news}         & --   & 82.9 \\ 
\bottomrule
\end{tabular}
\caption{Spanned non-EDU sentences, \%}
\label{tab:genre_edu_coverage}
\end{table}

While \cite{soricut-marcu-2003-sentence} briefly mention a 95\% coverage of sentences spanned by well-formed rhetorical subtrees in RST-DT, our analysis, based on automatic sentence segmentation and counting within binarized trees (the standard format for RST parsing), suggests a more conservative estimate of 86\%. Notably, even among non-elementary sentences (those containing at least two elementary units) there remains a prevalence of 79.4\% well-formed rhetorical trees in the corpus.

\section{RRT Preprocessing Details}
\label{sec:rrt_relations_mapper}

Table \ref{tab:rrt_renaming} provides information about the common renaming of mislabeled samples in RRT.

\begin{table}[h!]
\centering
\small
\begin{tabular}{@{}ll@{}}
\toprule
Original Annotation                                                                              & Preprocessing             \\ \midrule
antithesis  & Attribution  \\
cause, effect, cause-effect  & Cause-effect  \\
condition, motivation  & Condition  \\
\begin{tabular}[c]{@{}l@{}}evaluation, interpretation, \\ interpretation-evaluation\end{tabular} & Interpetation-evaluation \\ \midrule
\textsc{restatement\_SN}                                                                                  & \textsc{Condition\_SN}             \\
\textsc{restatement\_NS}                                                                                  & \textsc{Elaboration\_NS}           \\
\textsc{solutionhood\_NS}                                                                                 & \textsc{Solutionhood\_SN}          \\
\textsc{preparation\_NS}                                                                                  & \textsc{Elaboration\_NS}           \\
\textsc{elaboration\_SN}                                                                                  & \textsc{Preparation\_SN}           \\
\textsc{background\_NS}                                                                                   & \textsc{Elaboration\_SN}           \\ \bottomrule
\end{tabular}
\caption{Common renaming of mislabeled relations \newline during RRT preprocessing.}
\label{tab:rrt_renaming}
\end{table}

The mislabelings, which persist in version 2.1 and are consequently addressed during corpus preprocessing, can be attributed to the following factors:
\begin{itemize}
    \item{\textbf{Relation selection errors.} The Antithesis relation is intentionally excluded from the corpus during annotation. However, a few instances of this class within the corpus clearly imply the Attribution relation. Furthermore, Restatement\_SN(NS), Preparation\_NS, Elaboration\_SN are considered impossible according to the annotation manual.}
    \item{\textbf{Artifacts of shifting relation definitions.} In pursuit of objectivity and annotation agreement, \citet{pisarevskaya2017towards} combined or eliminated certain initial relations (cause, effect, motivation, evaluation, interpretation). Nevertheless, remnants of these fine-grained labels persist within the corpus.}
\end{itemize}

\section{RRG Construction Example}
\label{sec:construction_example}

\begin{figure}[ht!]
\centering
    \begin{subfigure}[b]{0.42\textwidth}
        \centering
        \includegraphics[width=0.9\textwidth]{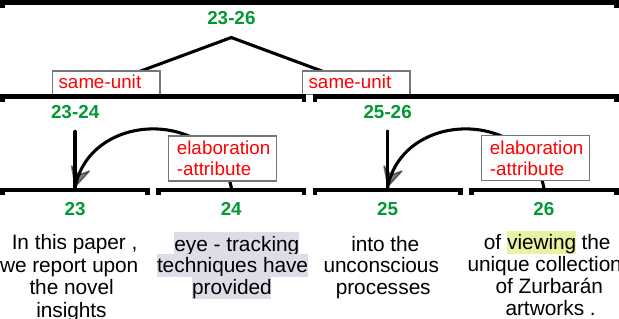}
        \par\bigskip
        \par
        \caption{GUM annotation.}
        \label{fig:zurbaran_en}
    \end{subfigure}
    \par\bigskip
    \begin{subfigure}[b]{0.47\textwidth}
        \centering
        \includegraphics[width=0.4\textwidth]{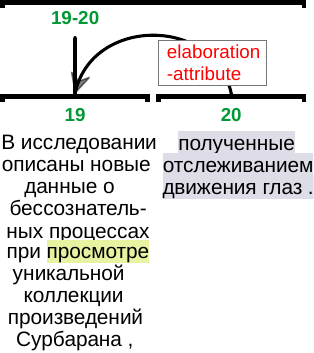}
        \caption{Corresponding RRG annotation. Literally: [In this paper are reported the novel insights into the unconcious processes of viewing the unique collection of Zurbarán works]$_{19}$ [provided by eye-tracking techniques.]$_{20}$}
        \label{fig:zurbaran_ru}
    \end{subfigure}
    
    \caption{Example of N-to-One EDU mapping. From~\texttt{academic\_art}.}
    \label{fig:n-to-one-mapping2}
\end{figure}

We use an additional example in Figure \ref{fig:n-to-one-mapping2} to illustrate the details of the RRG creation process described in section \ref{sec:rrg}.

\paragraph{Translation} The first step involves translating the English sentence presented in Figure~\ref{fig:zurbaran_en} into an academic Russian equivalent (Figure~\ref{fig:zurbaran_ru}).
Machine EDU-level translation,\footnote{DeepL is used for this example.} as employed in related work, yields an incomprehensible sequence of unrelated phrases: \textcyrillic{[В этой статье мы сообщаем о новых открытиях]$_{23}$  [методы слежения за глазами обеспечили]$_{24}$ [в бессознательные процессы]$_{25}$ [осмотр уникальной коллекции произведений искусства Зурбарана.]$_{26}$}. Manual translation, on the other hand, not only preserves coherence but also incorporates genre-specific adaptations to ensure alignment with established conventions of Russian academic writing. These adaptations include the use of academic clichés and the passive voice. Additionally, factual adaptations ensure accurate translations of terms and names, such as \textit{eye-tracking} to \textit{\textcyrillic{отслеживание движения глаз}}, and \textit{Zurbarán} to \textit{\textcyrillic{Сурбаран}}.

\paragraph{Rhetorical Structure Alignment} The order of EDUs differs between English and Russian. English EDUs 23, 25, and 26 combine into a single unit in Russian due to \textit{viewing} translating to the noun \textcyrillic{просмотр}. This collapses the \textsc{same-unit} relation, resulting in a direct alignment of the remaining \textsc{Elaboration\_NS}.

\section{RRG Polishing Details}
\label{sec:rurstg_details}

\begin{table}[!h]
\centering
\scriptsize
\begin{tabular}{@{}ll@{}}
\toprule
What                                                                                                                                    & How                      \\ \midrule
(original name form; years of birth and death)                                                                                         & joint-list               \\ \midrule
emojis separated from sentences                                                                                                         & evaluation-comment       \\ \midrule
\begin{tabular}[c]{@{}l@{}} "\textcyrillic{посвящённый} ..." (devoted), \\ "\textcyrillic{нацеленный} ..." (targeting), \\ and "\textcyrillic{направленный} ..." (aimed)\end{tabular} & purpose-attribute        \\ \midrule
"{[}\textcyrillic{также}{]} \textcyrillic{известный как}" ({[}also{]} known as)                                                                                       & restatement-partial      \\
"\textcyrillic{Как я} {[}\textcyrillic{уже}{]} \textcyrillic{говорил(а)}, ..." (As I said,)                                                                                          & organization-preparation \\ \bottomrule
\end{tabular}
\caption{Standardization of inconsistent annotations inherited from GUM$_{9.1}$.}
\label{tab:standardization_gum}
\end{table}

\section{Implementation Details}
\label{sec:implementation}

Table \ref{tab:hyperparameters} shows the hyperparameters used in our experiments. The experiments are performed on an NVIDIA Tesla v100 GPU. A single run takes 4 to 8 GPU hours, depending on the dataset and batch size.

\begin{table}[h!]
\centering
\scriptsize
\begin{tabular}{@{}lclll@{}}
\toprule
                                                                & \multicolumn{1}{l}{\textbf{RST-DT}} & \textbf{GUM} & \textbf{RRG} & \textbf{RRT} \\ \midrule
batch size (\# of trees)                                        & \multicolumn{1}{l}{2}               & 1            & 1            & 6            \\ %\midrule
$b_{\textrm{DWA}}$ (\# of trees)                                & \multicolumn{1}{l}{12}              & 12           & 12           & 24           \\ \midrule
\multicolumn{5}{c}{  \texttt{LM}  }                                                                                                                \\ \midrule
hidden size                                                     & \multicolumn{4}{c}{1024}                                                         \\ %\midrule
sliding window length                                           & \multicolumn{4}{c}{400}                                                          \\ %\midrule
learning rate                                                   & \multicolumn{4}{c}{2e-05}                                                        \\ \midrule
\multicolumn{5}{c}{  \texttt{Parser}  }                                                                                                            \\ \midrule
hidden size                                                     & \multicolumn{1}{l}{1024}            & 1024         & 1024         & 768          \\ %\midrule
dropout (segmenter input)                                       & \multicolumn{4}{c}{0.4}                                                          \\ %\midrule
dropout (encoder input)                                         & \multicolumn{4}{c}{0.5}                                                          \\ %\midrule
learning rate                                                   & \multicolumn{4}{c}{1e-04}                                                        \\ \midrule
\multicolumn{5}{c}{  \texttt{ToNy}  }                                                                                                              \\ \midrule
hidden size                                                     & \multicolumn{4}{c}{200}                                                          \\ \midrule
\multicolumn{5}{c}{  \texttt{E-BiLSTM}  }                                                                                                             \\ \midrule
hidden size                                                     & \multicolumn{4}{c}{512}                                                         \\ \bottomrule
\end{tabular}

\caption{Parameters used in the experiments.}
\label{tab:hyperparameters}
\end{table}

\section{Relation Classification Results}
\label{sec:clf_details}

\begin{table}[!h]
\centering
\scriptsize
\begin{tabular}{@{}lrrrr@{}}
\toprule
                              & \multicolumn{1}{l}{\textbf{P}} & \multicolumn{1}{l}{\textbf{R}} & \multicolumn{1}{l}{\textbf{F1}} & \multicolumn{1}{l}{\textbf{Num.}} \\ \midrule
\multicolumn{5}{c}{RRT}                                                                                                                                               \\ 
\midrule
Attribution\_NS     & 87.21 & 97.40 & 92.02 & 77    \\
Attribution\_SN     & 77.05 & 94.95 & 85.07 & 198   \\
Background\_SN      & 00.00 & 00.00 & 00.00 & 10    \\
Cause-effect\_NS    & 50.88 & 37.18 & 42.96 & 78    \\
Cause-effect\_SN    & 43.18 & 48.72 & 45.78 & 78    \\
Comparison\_NN      & 35.71 & 26.32 & 30.30 & 38    \\
Concession\_NS      & 83.33 & 90.91 & 86.96 & 22    \\
Concession\_SN      & 40.00 & 20.00 & 26.67 & 10    \\
Condition\_NS       & 53.47 & 75.00 & 62.43 & 72    \\
Condition\_SN       & 62.38 & 67.74 & 64.95 & 93    \\
Contrast\_NN        & 70.94 & 76.60 & 73.66 & 188   \\
Elaboration\_NS     & 52.72 & 71.21 & 60.59 & 639   \\
Evidence\_NS        & 26.67 & 08.89 & 13.33 & 45    \\
Evidence\_SN        & 00.00 & 00.00 & 00.00 & 12    \\
Interpretation-evaluation\_NS   & 45.24 & 39.58 & 42.22 & 144   \\
Interpretation-evaluation\_SN   & 33.33 & 15.38 & 21.05 & 13    \\
Joint\_NN           & 72.18 & 60.12 & 65.60 & 682   \\
Preparation\_SN     & 56.44 & 48.72 & 52.29 & 117   \\
Purpose\_NS         & 89.06 & 78.08 & 83.21 & 73    \\
Purpose\_SN         & 55.00 & 57.89 & 56.41 & 19    \\
Restatement\_NN     & 33.33 & 22.73 & 27.03 & 22    \\
Sequence\_NN        & 59.72 & 30.50 & 40.38 & 141   \\
Solutionhood\_SN    & 51.16 & 48.89 & 50.00 & 45    \\
same-unit\_NN       & 59.02 & 45.00 & 51.06 & 80    \\
\\
\textit{Macro avg.} & 51.58 & 48.41 & 48.92 & 2896  \\
\midrule
\multicolumn{5}{c}{RRG}                                                                                                                                               \\ \midrule
adversative\_NN & 24.32 & 17.31 & 20.22 &     52    \\
adversative\_NS & 35.85 & 33.33 & 34.55 &     57    \\
adversative\_SN & 36.23 & 51.02 & 42.37 &     49    \\
attribution\_NS & 84.00 & 72.41 & 77.78 &     29    \\
attribution\_SN & 69.47 & 88.35 & 77.78 &    103    \\
causal\_NS      & 29.55 & 16.46 & 21.14 &     79    \\
causal\_SN      & 07.14 & 05.88 & 06.45 &     17    \\
context\_NS     & 60.56 & 42.16 & 49.71 &    102    \\
context\_SN     & 35.24 & 30.58 & 32.74 &    121    \\
contingency\_NS & 71.43 & 71.43 & 71.43 &     14    \\
contingency\_SN & 86.49 & 84.21 & 85.33 &     38    \\
elaboration\_NS & 50.66 & 69.33 & 58.54 &    551    \\
evaluation\_NS  & 33.80 & 23.30 & 27.59 &    103    \\
evaluation\_SN  & 50.00 & 07.14 & 12.50 &     14    \\
explanation\_NS & 54.41 & 26.62 & 35.75 &    139    \\
explanation\_SN & 20.00 & 03.57 & 06.06 &     28    \\
joint\_NN       & 60.69 & 71.48 & 65.64 &    568    \\
mode\_NS        & 46.43 & 31.71 & 37.68 &     41    \\
mode\_SN        & 00.00 & 00.00 & 00.00 &      3    \\
organization\_NS    & 73.68 & 96.55 & 83.58 &     29    \\
organization\_SN    & 78.57 & 65.13 & 71.22 &    152    \\
purpose\_NS     & 85.07 & 82.61 & 83.82 &     69    \\
purpose\_SN     & 75.00 & 85.71 & 80.00 &      7    \\
restatement\_NN & 37.50 & 32.14 & 34.62 &     28    \\
restatement\_NS & 16.67 & 04.00 & 06.45 &     25    \\
same-unit\_NN   & 82.61 & 45.97 & 59.07 &    124    \\
topic\_SN       & 63.27 & 73.81 & 68.13 &     42    \\
\\

\textit{Macro avg.}       & 50.69   & 45.64 & 46.30 &   2584    \\ 
\bottomrule
\end{tabular}
\caption{Performance of the relation classification on Russian corpora.}
\label{tab:clf_results}
\end{table}

\begin{figure}[!h]
\centering
    \begin{subfigure}[b]{0.47\textwidth}
        % \centering
        \includegraphics[width=\textwidth]{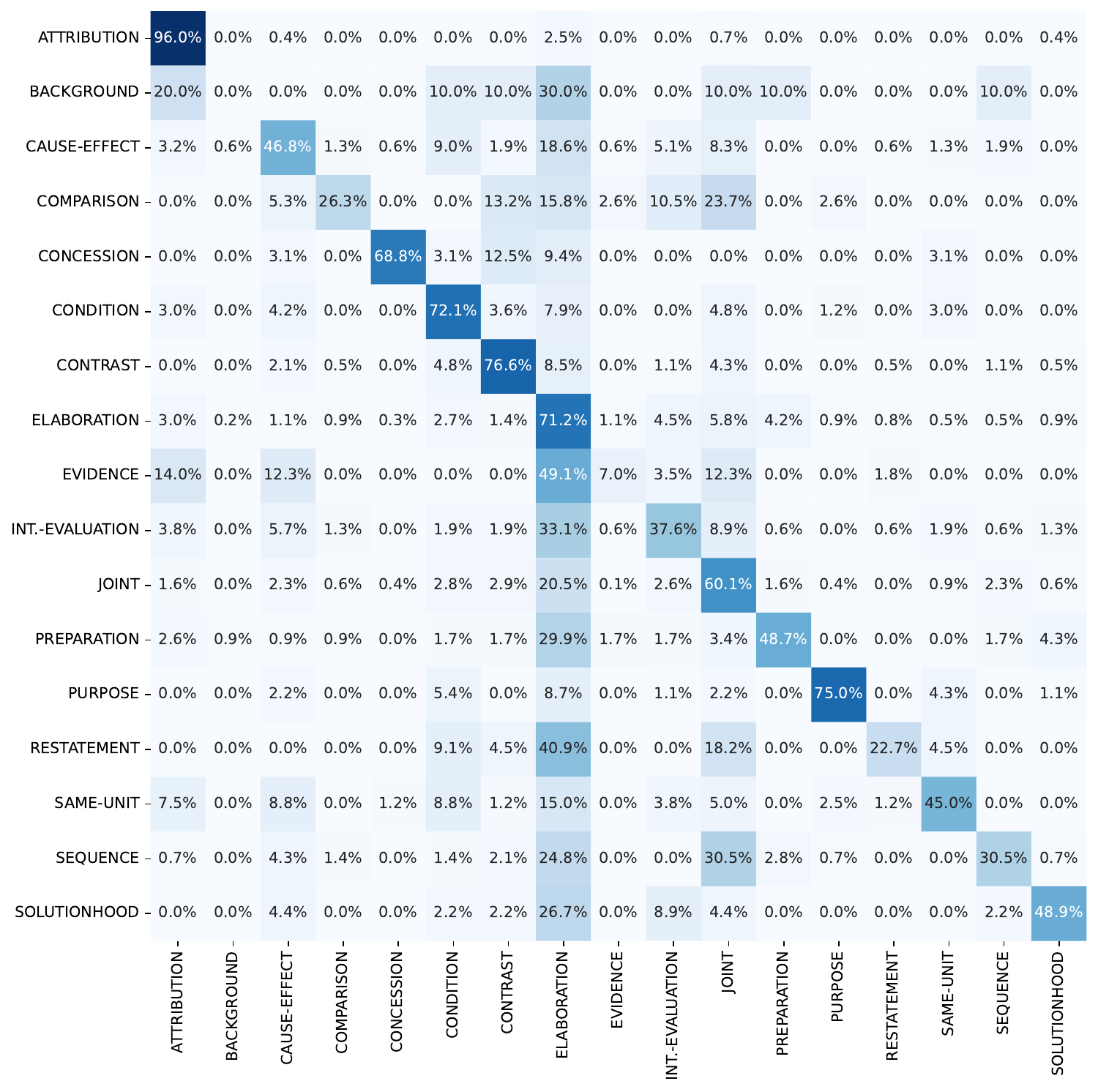}
        \caption{RRT}
        \label{fig:confusion_rrt}
    \end{subfigure}
    \par\bigskip
    % \hfill
    \begin{subfigure}[b]{0.47\textwidth}
        % \centering
        \includegraphics[width=\textwidth]{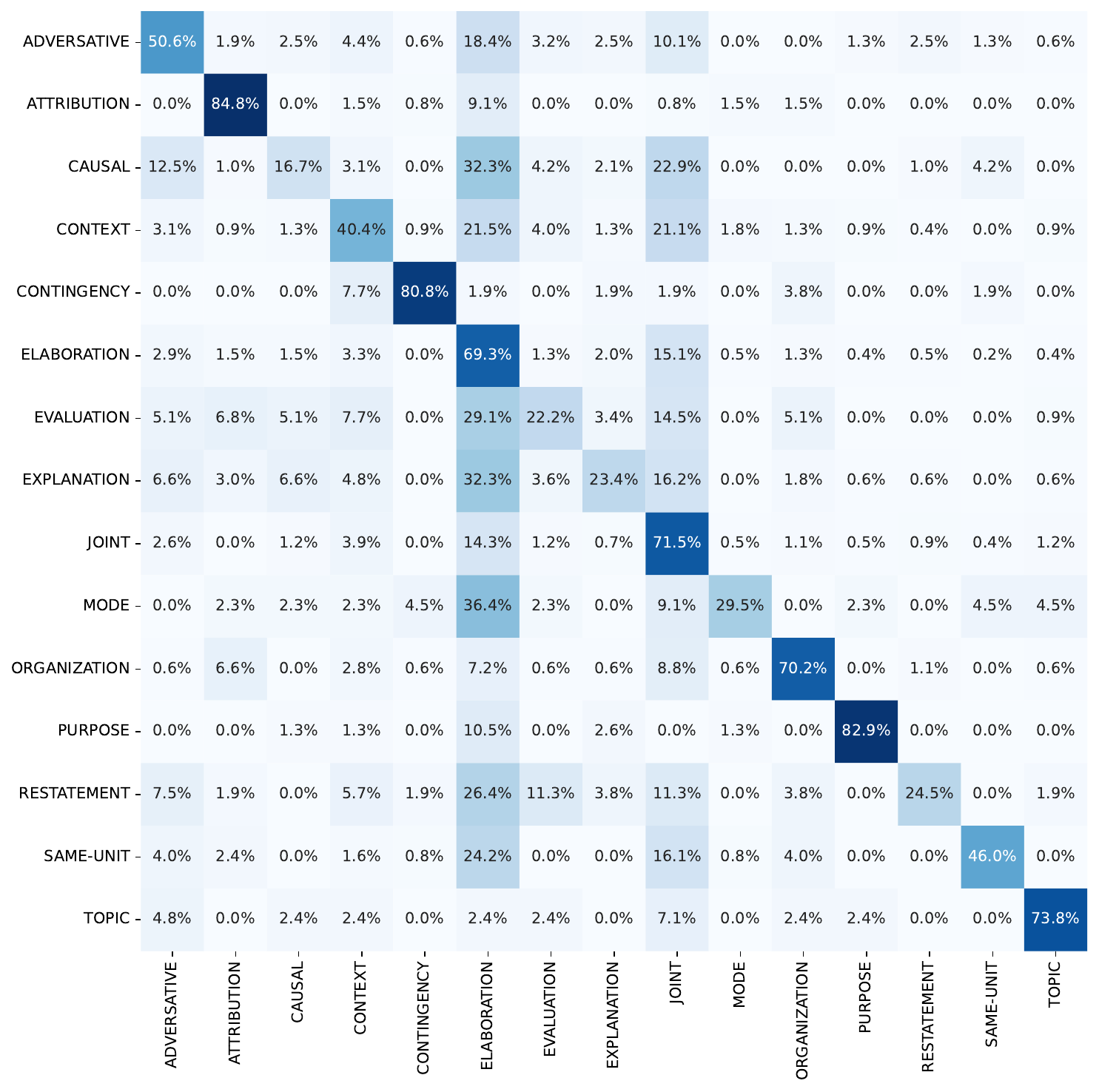}
        \caption{RRG}
        \label{fig:confusion_rrg}
    \end{subfigure}
    
    \caption{Confusion matrices for the relation classification on Russian corpora; nuclearity omitted.}
    \label{fig:relation_confusion}
\end{figure}

\begin{figure*}[!ht]
\centering
    \begin{subfigure}[b]{0.42\textwidth}
        \centering
        \includegraphics[width=\textwidth]{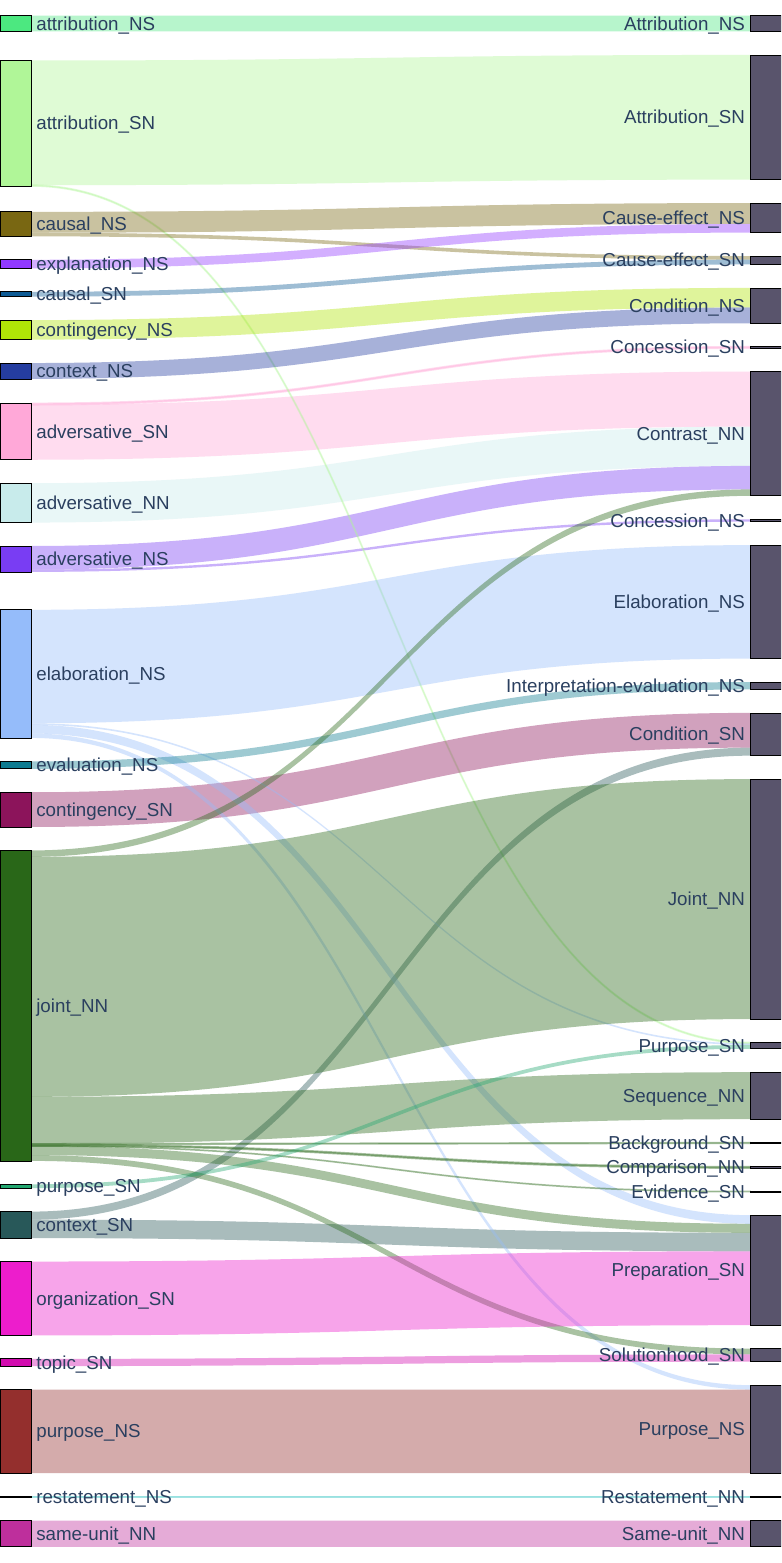}
        \caption{RRT Classifier $\rightarrow$ RRG}
        \label{fig:transfer_rrt-rrg}
    \end{subfigure}
    \hfill
    \begin{subfigure}[b]{0.42\textwidth}
        \centering
        \includegraphics[width=\textwidth]{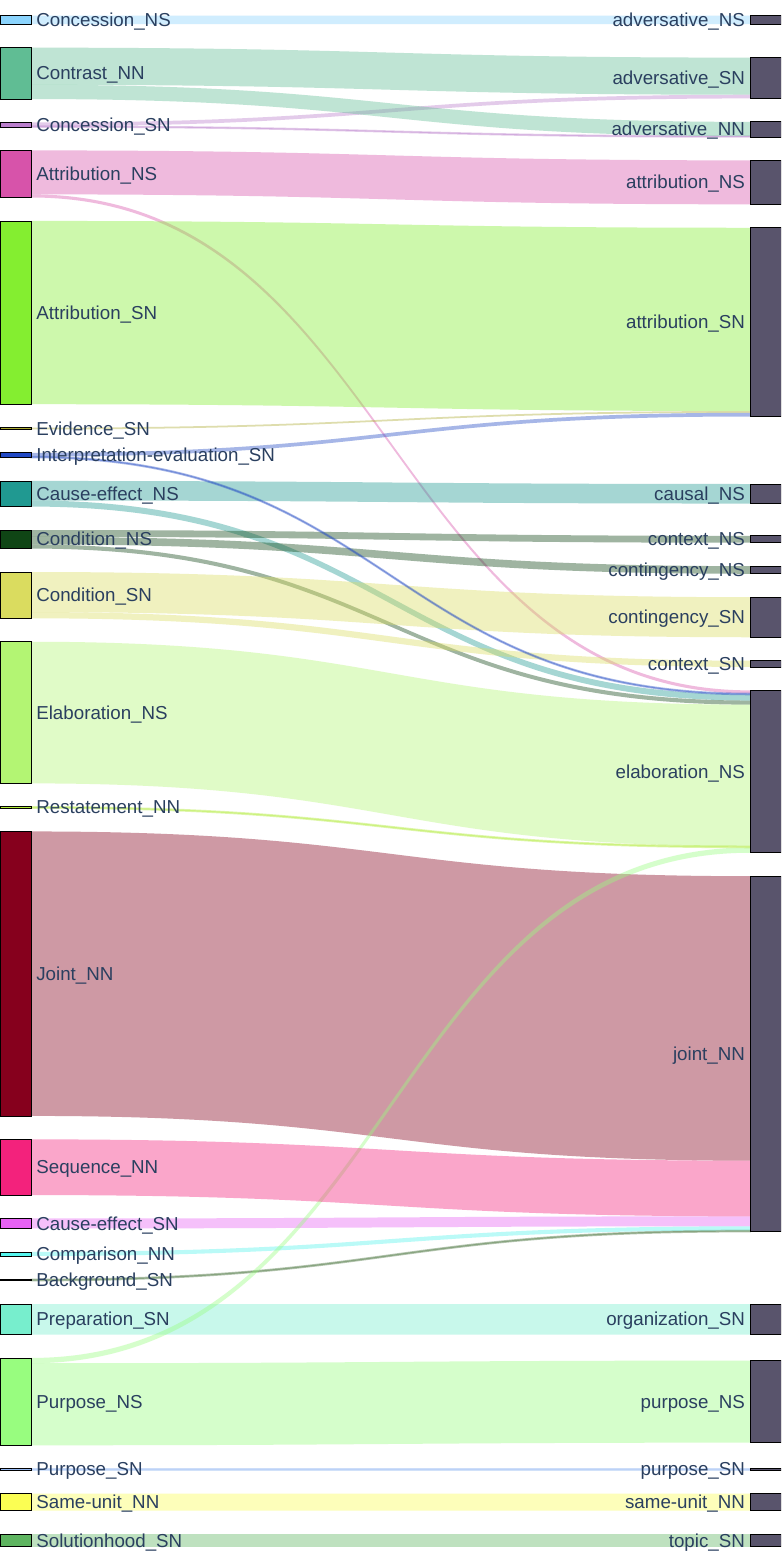}
        \caption{RRG Classifier $\rightarrow$ RRT}
        \label{fig:transfer_rrt-rrg}
    \end{subfigure}
    \par\bigskip

    \caption{A visual representation of the cross-dataset alignment between ground truth and predicted RST relations.}
 
    \label{fig:clf_transfer_results}
\end{figure*}

Table \ref{tab:clf_results} presents a detailed rhetorical relation classification performance for each corpus employing a standalone classifier. The task is treated in the context of the end-to-end system, with merged relation and nuclearity. Figure \ref{fig:relation_confusion} shows confusion matrices for the same classification models focusing only on the coarse-grained relation. Although the RRG-trained classifier achieved better performance for some mirroring relations (\textsc{Contingency/Condition}, \textsc{Purpose}, \textsc{Topic/Solutionhood}), it struggled with causal relations (16.7\% for RRG's \textsc{Causal} compared to 46.8\% for RRT's \textsc{Cause-Effect}). This can be attributed to the classifier's reliance on discourse cues, as only 23.6\% of DU pairs in RRG with a causal cue represent an actual causal relation, compared to 47.7\% in RRT. Notably, \textsc{Explanation} (13.9\%), \textsc{Elaboration} (11.6\%), \textsc{Joint} (10.0\%), and \textsc{Context} (8.4\%) are the most prevalent non-causal relations with causal markers in the RRG corpus.

Overlapping RST relation\_nuclearity classes across two corpora are illustrated in Figure \ref{fig:clf_transfer_results}. Confidently predicted relations (entropy >75th percentile) are shown on the right, with the target corpus's ground truth relations on the left. Only frequent transitions (>2.5\% of gold class) are included. These figures reveal recurring patterns of overlapping relations in the two annotation types. The classes \textsc{Organization\_NS}, \textsc{Mode}, \textsc{Context\_SN}, and \textsc{Organization\_NS} in the RRG corpus do not correspond with certain classes in RRT when examining the mentioned discourse unit features. The RRT-trained classifier consistently assigns the \textsc{Condition} class to both RRG's \textsc{contingency} (contingency-condition) and \textsc{context} (context-circumstance) classes. For parsing efficiency, RRG merges its specific adversative classes (antithesis, concession, contrast) into a single \textsc{adversative} category. This unified category maps to two distinct relations in the RRT: \textsc{Contrast} and \textsc{Concession}, leading to inconsistencies in nuclearity correspondence. The classifiers exhibit similar error patterns across both corpora. For instance, despite having its own dedicated Evidence relation within the broader \textsc{explanation} category, the RRG classifier consistently misidentifies the RRT's \textsc{Evidence} samples as \textsc{attribution}, mirroring 14\% of the RRT classifier's predictions. This suggests a bias in both models towards interpreting references to information sources as attributions, regardless of the intended meaning. Meanwhile, RRT's \textsc{Cause-effect} class absorbs \textsc{explanation}'s Justify and Motivation, encompassing both event causality and justifications (except for \textsc{Evidence}). 

\begin{figure}[!ht]
\centering
    \begin{subfigure}[b]{0.47\textwidth}
        \centering
        \includegraphics[width=0.8\textwidth]{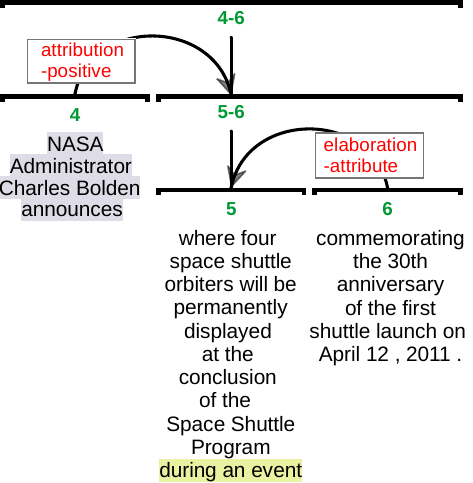}
        \caption{Original annotation from GUM$_{9.1}$.}
        \label{fig:transfer_rrt-rrg}
    \end{subfigure}
    \par\bigskip
    \begin{subfigure}[b]{0.47\textwidth}
        \centering
        \includegraphics[width=\textwidth]{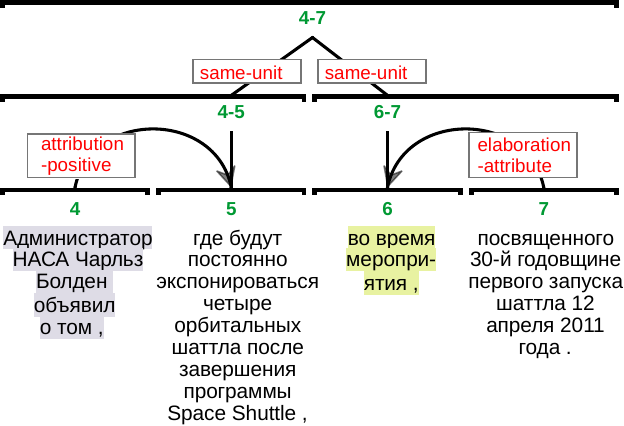}
        \caption{RRG corpus annotation. Commas mark EDU boundaries as follows: [NASA Administrator Charles Bolden announces]$_{4}$ [where four space shuttle orbiters will be permanently displayed at the conclusion of the Space Shuttle Program]$_{5}$ [during an event]$_{6}$ [commemorating the 30th anniversary of the first shuttle launch on April 12, 2011.]$_{7}$.}
        \label{fig:transfer_rrt-rrg}
    \end{subfigure}
    \par\bigskip
    \begin{subfigure}[b]{0.47\textwidth}
        \centering
        \includegraphics[width=0.25\textwidth]{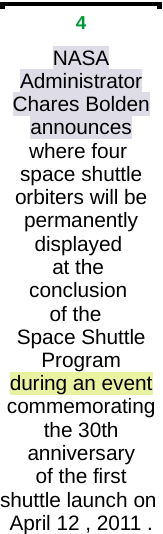}
        \caption{RRG parser prediction for English text.}
        \label{fig:transfer_rrt-rrg}
    \end{subfigure}

    \caption{An example of the zero-shot cross-language segmentation errors. From \texttt{GUM\_news\_nasa}.}
 
    \label{fig:clf_transfer_commas}
\end{figure}

\section{Genre-wise Evaluation}
\label{sec:genrewise_evaluation}

Tables \ref{tab:monolingual_all}, \ref{tab:zeroshot_all}, and \ref{tab:bilingual_all} provide detailed performance metrics for the end-to-end RST parsing in both languages. The monolingual Russian parser, when applied to English text in the zero-shot setting (Table~\ref{tab:zeroshot_all}), exhibits segmentation errors illustrated in Figure~\ref{fig:clf_transfer_commas}.

\begin{table*}[h]
\centering
\scriptsize
\begin{tabular}{@{}llllll|lllll@{}}
\toprule
             & \multicolumn{5}{c}{\textbf{en}}                                 & \multicolumn{5}{c}{\textbf{ru}}                                 \\ \midrule
             & Segm       & S          & N          & R          & Full       & Segm       & S          & N          & R          & Full       \\ \midrule
academic     & 94.6 ± 0.6 & 72.7 ± 1.3 & 64.0 ± 1.9 & 56.9 ± 1.7 & 56.3 ± 1.7 & 94.6 ± 0.5 & 72.6 ± 1.8 & 62.9 ± 1.1 & 55.8 ± 0.8 & 55.7 ± 0.8 \\
bio          & 97.7 ± 0.6 & 68.1 ± 1.8 & 57.0 ± 2.9 & 53.2 ± 2.1 & 51.5 ± 2.1 & 98.5 ± 0.3 & 69.0 ± 1.8 & 58.4 ± 1.1 & 52.8 ± 1.2 & 52.2 ± 1.2 \\
conversation & 95.5 ± 0.3 & 49.5 ± 1.3 & 39.0 ± 1.5 & 29.8 ± 1.4 & 29.3 ± 1.6 & 95.5 ± 0.5 & 48.5 ± 1.2 & 33.8 ± 1.1 & 27.4 ± 1.2 & 25.9 ± 1.4 \\
fiction      & 93.9 ± 0.7 & 59.3 ± 2.4 & 47.8 ± 2.9 & 39.7 ± 2.2 & 38.5 ± 2.3 & 96.2 ± 0.6 & 61.0 ± 1.1 & 47.3 ± 1.2 & 38.2 ± 0.6 & 36.7 ± 1.0 \\
interview    & 95.1 ± 0.4 & 73.8 ± 0.6 & 65.7 ± 1.3 & 55.3 ± 1.2 & 55.1 ± 1.1 & 96.6 ± 0.3 & 71.6 ± 1.9 & 60.3 ± 0.7 & 47.3 ± 0.8 & 47.3 ± 0.8 \\
news         & 94.6 ± 0.8 & 69.0 ± 1.9 & 60.4 ± 2.4 & 56.7 ± 2.0 & 55.0 ± 2.1 & 96.3 ± 0.5 & 65.7 ± 2.1 & 54.2 ± 3.2 & 47.6 ± 1.2 & 45.9 ± 1.7 \\
reddit       & 93.3 ± 0.6 & 60.5 ± 1.1 & 51.5 ± 1.4 & 44.5 ± 1.6 & 44.0 ± 1.4 & 97.7 ± 0.3 & 61.1 ± 1.3 & 48.6 ± 1.6 & 42.7 ± 1.4 & 41.5 ± 1.7 \\
speech       & 97.5 ± 0.4 & 79.1 ± 1.7 & 67.4 ± 2.4 & 57.8 ± 1.8 & 57.6 ± 2.0 & 96.0 ± 0.6 & 70.5 ± 2.5 & 58.7 ± 1.9 & 50.9 ± 0.5 & 50.2 ± 0.5 \\
textbook     & 97.5 ± 0.3 & 78.7 ± 1.3 & 66.1 ± 1.8 & 57.4 ± 2.0 & 57.0 ± 1.9 & 97.4 ± 0.3 & 76.0 ± 2.0 & 62.7 ± 2.3 & 54.6 ± 2.0 & 53.6 ± 2.0 \\
vlog         & 95.6 ± 0.5 & 61.9 ± 1.0 & 48.8 ± 2.0 & 43.5 ± 1.5 & 41.7 ± 1.7 & 97.9 ± 0.3 & 65.8 ± 2.1 & 43.2 ± 2.1 & 38.8 ± 1.5 & 35.5 ± 1.3 \\
voyage       & 94.6 ± 0.5 & 67.2 ± 1.9 & 51.6 ± 2.2 & 44.6 ± 2.0 & 44.1 ± 1.9 & 99.0 ± 0.1 & 73.7 ± 0.9 & 58.1 ± 0.8 & 50.4 ± 1.2 & 49.3 ± 1.0 \\
whow         & 97.3 ± 0.3 & 75.7 ± 0.9 & 64.3 ± 1.9 & 58.6 ± 1.7 & 57.0 ± 1.7 & 97.8 ± 0.5 & 75.5 ± 1.7 & 64.4 ± 2.3 & 55.5 ± 2.1 & 54.1 ± 2.2 \\
             &            &            &            &            &            &            &            &            &            &            \\
\textit{all} & 95.5 ± 0.1 & 66.9 ± 0.5 & 56.1 ± 0.3 & 48.8 ± 0.4 & 47.9 ± 0.4 & 96.9 ± 0.2 & 66.5 ± 0.4 & 53.3 ± 0.6 & 45.8 ± 0.5 & 44.6 ± 0.4 \\ \bottomrule
\end{tabular}
\caption{Detailed evaluation of the monolingual parsers.}
\label{tab:monolingual_all}
\end{table*}

\begin{table*}[ht]
\centering
\scriptsize
\begin{tabular}{@{}llllll|lllll@{}}
\toprule
             & \multicolumn{5}{c}{\textbf{ru} $\rightarrow$ en}                                      & \multicolumn{5}{c}{\textbf{en} $\rightarrow$ ru}                                      \\ \midrule
             & Segm       & S          & N          & R          & Full       & Segm       & S          & N          & R          & Full       \\ \midrule
\textit{academic}     & 83.1 ± 1.3 & 52.0 ± 4.3 & 43.2 ± 3.2 & 39.0 ± 3.0 & 38.7 ± 2.9 & 93.1 ± 0.9 & 69.2 ± 0.8 & 61.5 ± 0.2 & 52.1 ± 0.9 & 52.1 ± 0.9 \\
\textit{bio}          & 94.4 ± 0.5 & 63.0 ± 1.8 & 50.1 ± 2.8 & 45.9 ± 3.0 & 44.8 ± 3.2 & 97.3 ± 0.4 & 66.3 ± 1.0 & 54.6 ± 0.5 & 47.5 ± 0.8 & 46.3 ± 0.9 \\
\textit{conversation} & 91.6 ± 0.6 & 42.4 ± 1.7 & 30.8 ± 2.0 & 23.5 ± 1.2 & 22.8 ± 1.4 & 94.4 ± 0.7 & 45.5 ± 2.5 & 32.9 ± 3.3 & 23.2 ± 2.5 & 22.1 ± 2.4 \\
\textit{fiction}      & 85.3 ± 0.8 & 47.8 ± 2.6 & 35.9 ± 2.6 & 28.8 ± 1.9 & 27.7 ± 1.7 & 94.9 ± 0.7 & 60.0 ± 2.4 & 48.1 ± 2.8 & 38.1 ± 1.8 & 37.2 ± 1.7 \\
\textit{interview}    & 83.2 ± 1.4 & 43.9 ± 3.6 & 37.1 ± 2.3 & 29.6 ± 2.9 & 29.5 ± 2.7 & 95.6 ± 0.8 & 69.7 ± 1.3 & 58.2 ± 1.0 & 46.9 ± 0.8 & 46.1 ± 1.0 \\
\textit{news}         & 84.5 ± 1.8 & 45.9 ± 3.3 & 38.7 ± 3.4 & 36.9 ± 2.9 & 34.8 ± 2.6 & 93.5 ± 1.2 & 61.9 ± 1.2 & 51.8 ± 1.7 & 45.8 ± 0.9 & 44.4 ± 1.0 \\
\textit{reddit}       & 83.1 ± 1.4 & 37.1 ± 2.7 & 30.7 ± 1.8 & 24.9 ± 2.5 & 24.6 ± 2.3 & 97.1 ± 0.4 & 59.8 ± 2.1 & 48.5 ± 1.7 & 41.1 ± 0.8 & 40.6 ± 0.8 \\
\textit{speech}       & 83.7 ± 1.6 & 44.6 ± 2.1 & 34.8 ± 1.3 & 29.8 ± 2.4 & 29.5 ± 2.5 & 94.5 ± 0.5 & 69.8 ± 1.1 & 56.6 ± 0.9 & 48.6 ± 1.6 & 47.8 ± 1.3 \\
\textit{textbook}     & 87.8 ± 1.4 & 56.2 ± 2.3 & 45.7 ± 2.3 & 39.9 ± 2.3 & 39.2 ± 2.1 & 95.1 ± 0.3 & 71.2 ± 0.9 & 58.0 ± 1.8 & 51.9 ± 0.6 & 51.4 ± 0.6 \\
\textit{vlog}         & 88.1 ± 1.9 & 52.7 ± 3.4 & 35.7 ± 3.1 & 32.8 ± 3.5 & 30.2 ± 3.9 & 97.2 ± 0.1 & 61.6 ± 1.8 & 41.5 ± 0.6 & 36.1 ± 1.0 & 33.3 ± 0.7 \\
\textit{voyage}       & 85.1 ± 1.2 & 46.6 ± 2.6 & 34.9 ± 2.3 & 28.8 ± 1.7 & 28.7 ± 1.5 & 96.7 ± 0.3 & 71.6 ± 1.4 & 55.2 ± 1.6 & 48.8 ± 2.4 & 46.8 ± 1.9 \\
\textit{whow}         & 90.6 ± 1.8 & 58.7 ± 3.8 & 49.8 ± 3.9 & 42.9 ± 3.2 & 42.1 ± 3.0 & 96.5 ± 0.5 & 74.0 ± 1.6 & 61.9 ± 1.8 & 54.1 ± 1.7 & 52.0 ± 1.9 \\
\\
\textit{all}          & 86.9 ± 1.0 & 49.0 ± 2.2 & 38.6 ± 2.1 & 33.1 ± 1.9 & 32.2 ± 1.9 & 95.5 ± 0.3 & 63.9 ± 0.7 & 51.4 ± 1.0 & 43.4 ± 0.6 & 42.2 ± 0.6 \\ \bottomrule
\end{tabular}
\caption{Evaluating monolingual parsing transfer to a second language.}
\label{tab:zeroshot_all}
\end{table*}

\begin{table*}[ht]
\centering
\scriptsize
\begin{tabular}{@{}llllll|lllll@{}}
\toprule
             & \multicolumn{5}{c}{\textbf{en+ru} $\rightarrow$ en}                                 & \multicolumn{5}{c}{\textbf{en+ru} $\rightarrow$ ru}                                 \\ \midrule
             & Segm       & S          & N          & R          & Full       & Segm       & S          & N          & R          & Full       \\ \midrule
academic     & 94.2 ± 0.4 & 71.6 ± 1.1 & 63.1 ± 2.0 & 55.9 ± 2.1 & 55.5 ± 2.3 & 94.9 ± 0.6 & 72.9 ± 1.7 & 63.2 ± 1.6 & 55.3 ± 1.0 & 55.2 ± 1.0 \\
bio          & 97.6 ± 0.3 & 70.0 ± 0.9 & 58.4 ± 1.0 & 54.0 ± 1.4 & 52.5 ± 1.5 & 98.4 ± 0.4 & 68.1 ± 1.9 & 57.5 ± 1.7 & 51.4 ± 1.4 & 50.3 ± 1.4 \\
conversation & 95.1 ± 0.1 & 51.5 ± 1.5 & 39.2 ± 0.7 & 31.1 ± 1.4 & 30.2 ± 1.3 & 95.3 ± 0.4 & 47.8 ± 1.0 & 34.8 ± 1.3 & 28.9 ± 0.5 & 27.4 ± 0.5 \\
fiction      & 93.3 ± 0.6 & 59.2 ± 2.8 & 48.8 ± 2.3 & 41.2 ± 1.8 & 40.2 ± 1.8 & 96.6 ± 0.3 & 62.8 ± 1.9 & 49.6 ± 0.7 & 39.2 ± 2.0 & 38.0 ± 2.2 \\
interview    & 94.6 ± 0.5 & 71.7 ± 1.2 & 63.5 ± 1.8 & 55.2 ± 1.3 & 54.7 ± 1.2 & 96.9 ± 0.1 & 70.0 ± 1.7 & 60.2 ± 1.9 & 49.2 ± 1.8 & 48.8 ± 1.8 \\
news         & 94.8 ± 0.7 & 67.5 ± 2.4 & 59.2 ± 1.8 & 54.5 ± 1.6 & 52.9 ± 1.7 & 96.8 ± 0.7 & 68.5 ± 0.6 & 56.8 ± 1.7 & 49.6 ± 1.0 & 47.9 ± 1.4 \\
reddit       & 92.6 ± 0.8 & 58.5 ± 1.5 & 48.9 ± 2.3 & 43.0 ± 2.2 & 42.3 ± 2.2 & 97.2 ± 0.3 & 60.9 ± 1.6 & 49.4 ± 2.0 & 42.5 ± 1.6 & 41.7 ± 1.7 \\
speech       & 97.3 ± 0.3 & 75.7 ± 1.6 & 64.8 ± 1.9 & 57.2 ± 1.1 & 57.2 ± 1.1 & 96.3 ± 0.5 & 69.9 ± 2.4 & 57.5 ± 1.0 & 50.7 ± 1.1 & 50.1 ± 1.1 \\
textbook     & 97.5 ± 0.4 & 77.3 ± 1.7 & 65.3 ± 2.0 & 57.3 ± 0.8 & 56.4 ± 0.9 & 97.1 ± 0.3 & 77.1 ± 0.6 & 64.6 ± 1.0 & 56.1 ± 1.3 & 55.3 ± 1.1 \\
vlog         & 95.9 ± 0.4 & 62.8 ± 2.0 & 46.1 ± 2.6 & 42.8 ± 2.8 & 40.6 ± 2.7 & 97.8 ± 0.5 & 66.0 ± 1.7 & 46.0 ± 3.1 & 39.8 ± 3.4 & 36.5 ± 3.0 \\
voyage       & 94.2 ± 0.5 & 65.7 ± 2.5 & 49.5 ± 3.0 & 43.7 ± 2.6 & 43.4 ± 2.6 & 98.5 ± 0.3 & 76.4 ± 1.5 & 60.0 ± 1.9 & 51.7 ± 1.5 & 51.0 ± 1.4 \\
whow         & 97.2 ± 0.3 & 75.5 ± 1.3 & 65.0 ± 1.8 & 58.3 ± 1.9 & 56.8 ± 1.6 & 97.8 ± 0.3 & 75.9 ± 1.5 & 64.5 ± 2.5 & 56.3 ± 1.1 & 54.7 ± 1.5 \\
             &            &            &            &            &            &            &            &            &            &            \\
\textit{all} & 95.3 ± 0.1 & 66.4 ± 0.7 & 55.2 ± 0.6 & 48.6 ± 0.6 & 47.6 ± 0.7 & 96.8 ± 0.1 & 66.9 ± 0.4 & 54.3 ± 0.3 & 46.5 ± 0.4 & 45.4 ± 0.4 \\ \bottomrule
\end{tabular}
\caption{Bilingual parser performance.}
\label{tab:bilingual_all}
\end{table*}

\end{document}